\documentclass{article}

     \PassOptionsToPackage{square,sort,comma,numbers}{natbib}
     \usepackage[preprint]{neurips_2019}

\usepackage[utf8]{inputenc} % allow utf-8 input
\usepackage[T1]{fontenc}    % use 8-bit T1 fonts
\usepackage{hyperref}       % hyperlinks
\usepackage{url}            % simple URL typesetting
\usepackage{booktabs}       % professional-quality tables
\usepackage{amsfonts}       % blackboard math symbols
\usepackage{nicefrac}       % compact symbols for 1/2, etc.
\usepackage{microtype}      % microtypography

\usepackage{algorithm}
\usepackage{algorithmicx}
\usepackage{algpseudocode}
\usepackage{amsmath}
\usepackage{graphicx}

\usepackage{subfigure}

\newtheorem{myLemma}{Lemma}
\newtheorem{myDefinition}{Definition}

\title{Disentangling Dynamics and Returns: Value Function Decomposition with Future Prediction}

\author{%
  Hongyao Tang\textsuperscript{1},
  Jianye Hao\textsuperscript{1},
  Guangyong Chen\textsuperscript{2},
  Pengfei Chen\textsuperscript{3},
  Zhaopeng Meng\textsuperscript{1}, \\
  \textbf{Yaodong Yang\textsuperscript{1},
  Li Wang\textsuperscript{1}} \\
  \textsuperscript{1}College of Intelligence and Computing, Tianjin University,
  \textsuperscript{2}Tencent \\
  \textsuperscript{3}Department of Computers Science and Engineering, The Chinese University of Hong Kong \\
  \textsuperscript{1}\texttt{\{bluecontra,jianye.hao,mengzp,wangli\}@tju.edu.cn}, \textsuperscript{1}\texttt{yydapple@gmail.com} \\
  \textsuperscript{2}\texttt{gycchen@tencent.com}, \textsuperscript{3}\texttt{chenpf.cuhk@gmail.com}
}

\begin{document}

\maketitle

\begin{abstract}
  Value functions are crucial for model-free Reinforcement Learning (RL) to obtain a policy implicitly or guide the policy updates.
  %Value estimation is heavily influenced by the stochasticity of environmental dynamics and the quality of reward signals.
  Value estimation heavily depends on the stochasticity of environmental dynamics and the quality of reward signals.
  %Value estimation heavily depends on the quality of reward signals which are usually flawed or delayed in many practical problems.
  In this paper, we propose a two-step understanding of value estimation from the perspective of future prediction,
  through decomposing the value function into a reward-independent future dynamics part and a policy-independent trajectory return part.
  %through re-writing the value function into a composite form of a reward-independent part and a policy-independent part.
  %This induces a way to disentangle dynamics and returns during the value estimation process.
  %inducing a way to disentangle dynamics and returns based on a two-step understanding of the value estimation process in model-free RL.
  We then derive a practical deep RL algorithm from the above decomposition, consisting of a convolutional trajectory representation model, a conditional variational dynamics model to predict the expected representation of future trajectory and a convex trajectory return model that maps a trajectory representation to its return.
  %predicted future to its benefits.
  %a conditional variational dynamics model and a convolutional future trajectory return model.
  Our algorithm is evaluated in MuJoCo continuous control tasks and shows superior results under both common settings and delayed reward settings.
\end{abstract}

\section{Introduction}
Reinforcement learning (RL) is one promising approach to obtain the optimal policy in sequential decision-making problems based on reward signals during interaction with the environment.
Most popular RL algorithms are model-free since they do not need the access to the environment models.
%Model-free RL approaches do not need the access to the environment models thus are more popular and practical for real-world tasks.
Value functions play an important role in model-free RL \cite{Sutton1988ReinforcementLA}, which are usually used to derive a policy implicitly in value-based methods \cite{Mnih2015DQN} or guide the policy updates in policy-based methods \cite{Schulman2015TRPO, Silver2014DPG}.
With deep neural networks, value functions can be well-estimated even with large state and action space, making it practical for model-free RL to deal with more challenging tasks \cite{Lillicrap2015DDPG,Mnih2015DQN,Silver2016Go}.
%a lot of great success has been achieved in recent years, e.g., [xxxxxxxx].

Value functions define the expected cumulative rewards of a policy,
indicating the degree a state or an action could be beneficial.
%Value functions are usually estimated through Monte Carlo (MC) or Temporal Difference (TD) algorithms \cite{Sutton1988ReinforcementLA} with the ex entanglement of reward signals and environmental dynamics.
They are usually estimated through Monte Carlo (MC) or Temporal Difference (TD) algorithms \cite{Sutton1988ReinforcementLA},
without explicitly handling the entanglement of reward signals and environmental dynamics.
However, in practical problems,
the quality of value estimation is heavily crippled by highly stochastic dynamics and flawed or delayed reward.
%the entanglement of highly stochastic dynamics and flawed or delayed reward makes value estimation even intractable.
%Intuitively, instead of predicting the benefits directly, a policy is usually evaluated by human in a two-step way:
Intuitively, in contrast to the coupling manner, human beings usually evaluate a policy in a two-step way:
1) think how the environment would change afterwards; 2) then evaluate how good the predicted future could be.
Such a idea of future prediction is also proposed in cognitive behavior and neuroscience studies \cite{Atance2001EpisodicFT,Schacter2007RememberingTP,Schacter2007TheCN}.

Following this inspiration, in this paper, we look into the value function and re-write it as a composite form of:
%we re-write the value function and obtain a lower bound of it in a composite form of:
1) a reward-independent predictive dynamics function, which defines the expected representation of future state-action trajectory;
and 2) a policy-independent trajectory return function that maps any trajectory (representation) to its discounted cumulative reward.
This induces a two-step understanding of the value estimation process in model-free RL and provides a way to disentangle the dynamics and returns accordingly.
Further, we use modern deep learning techniques to build a practical algorithm
%called Value Decomposition with Future Prediction (VDFP)
based on the above decomposition, including a convolutional trajectory representation model, a conditional variational dynamics model and a convex trajectory return model.
%We evaluate our algorithm in MuJoCo continuous control tasks \cite{Brockman2016Gym,Todorov2012MuJoCo} and open source codes are available at [githublink].
%Our algorithm achieves state-of-the-art performance in MuJoCo continuous control task \cite{Brockman2016Gym,Todorov2012MuJoCo} and shows significant effectiveness and robustness under the challenging delayed reward settings.

Key contributions of this work are summarized as follows:

\begin{itemize}

\item We provide a new understanding of the value estimation process,
in a form of the explicit two-step composition between future dynamics prediction and trajectory return estimation.
%We provide a new understanding of the value function from the perspective of future prediction,
%in a composite form of future dynamics function and trajectory return function.
%allowing to disentangle the dynamics and returns for more effective and flexible value estimation in various problems.

\item We propose a decoupling way to learn value functions.
Through training the reward-independent predictive dynamics model and the policy-independent trajectory return model separately,
the value estimation process can be performed more effectively and flexibly in more challenging settings, e.g., delayed reward problems.

\item We propose a conditional Variational Auto-Encoder (VAE) \cite{Higgins2017Beta,Kingma2013AEVB} to model the underlying distribution of future trajectory representation.
Moreover, we use the generation process of the decoder as the predictive dynamics model and clip the generative noise with small variance to model the expectation of trajectory representation.

\item For reproducibility,
we conduct experiments on commonly adopted MuJoCo continuous control tasks \cite{Brockman2016Gym,Todorov2012MuJoCo} and perform ablation studies across each contribution.
Our algorithm achieves state-of-the-art performance under common settings and shows significant effectiveness and robustness under challenging delayed reward settings.
%Open source codes can be found in the Supplementary Material and will be released on GitHub soon.

\end{itemize}

\section{Background}
Consider a Markov Decision Process (MDP) $\left< \mathcal{S}, \mathcal{A}, \mathcal{P}, \mathcal{R}, \rho_0, \gamma, T\right>$,
defined with a set of states $\mathcal{S}$, a set of actions $\mathcal{A}$, the transition function $\mathcal{P}: \mathcal{S} \times \mathcal{A} \times \mathcal{S} \rightarrow \mathbb{R}_{\in [0,1]}$, the reward function $\mathcal{R}: \mathcal{S} \times \mathcal{A} \rightarrow \mathbb{R}$, the initial state distribution $\rho_0: \mathcal{S} \rightarrow \mathbb{R}_{\in [0,1]}$, the discounted factor $\gamma \in [0,1]$, and the finite horizon $T$.
An agent interacts with the MDP at discrete time steps by performing its policy $\pi: \mathcal{S} \rightarrow \mathcal{A}$,
%$\pi: \mathcal{S} \times \mathcal{A} \rightarrow \mathcal{R}_{\in [0,1]}$,
generating a trajectory of states and actions,
$\tau = \tau_{0:T} = \left(s_0, a_0, \dots, s_T , a_T \right)$, where $s_0 \sim \rho_0(s_0)$, $a_t \sim \pi(s_t)$ and $s_{t+1} \sim \mathcal{P}(s_{t+1}|s_t, a_t)$.
%The return is defined as the discounted sum of rewards over the trajectory $\tau$,
%$\R_{\tau} = \$
The objective of the agent is to
%obtain a policy that
maximize the expected cumulative discounted reward, denoted by $J(\pi) = \mathbb{E} \left[\sum_{t=0}^{T}\gamma^{t} r_t| \pi \right]$ where $r_t = \mathcal{R}(s_t,a_t)$.

In reinforcement learning,
%value functions are used to obtain the policy implicitly for value-based methods [dqn] or guide the gradient ascent of policy updates for policy-based methods [pg].
the state-action value function $Q$ is defined as the expected cumulative discounted reward for selecting action $a$ in state $s$, then following a policy $\pi$ afterwards:
\begin{equation}
\label{eqation:1}
    Q^{\pi}(s,a) = \mathbb{E} \left[\sum_{t=0}^{T}\gamma^{t} r_t|s_0=s, a_0=a; \pi \right]. % {\rm and} \ r_t \sim r(s_t, a_t).
\end{equation}
Similarly, the state value function $V$ denotes the expected cumulative discounted reward for performing a policy $\pi$ from a certain state $s$, i.e., $V^{\pi}(s) = \mathbb{E} \left[\sum_{t=0}^{T}\gamma^{t} r_t|s_0=s; \pi \right]$.
% Value functions are usually estimated with Monte Carlo or Temporal Difference algorithms [sutton].

For continuous control, a parameterized policy $\pi_{\theta}$, with parameters $\theta$, can be updated by taking the gradient of the objective $\nabla_{\theta} J(\pi_{\theta})$.
In actor-critic methods, the policy, known as the actor can be updated with the deterministic policy gradient theorem \cite{Silver2014DPG}:
\begin{equation}
\label{eqation:2}
   \nabla_{\theta} J(\pi_{\theta}) = \mathbb{E}_{s \sim \rho^{\pi}} \left[ \nabla_{\theta} \pi_{\theta}(s) \nabla_{a} Q^{\pi}(s,a)|_{a=\pi_{\theta}(s)}\right],
\end{equation}
where $\rho^{\pi}$ is the discounted state distribution under policy $\pi$.
The $Q$-function, also known as the critic, is usually estimated with Monte Carlo (MC) or Temporal Difference (TD) algorithms \cite{Sutton1988ReinforcementLA}.

\section{Model}
\label{section:model}
%The value function takes the form of the expectation of cumulative discounted reward % with respect to the policy.
%with no explicit concern on how the environment would change afterwards.
%It indicates the degree an action or a state could be beneficial
% indicating how good an action or a state is under a policy.
%while gives no explicit evidence on how the environment would change afterwards.
Value estimation faces the coupling of environmental dynamics and reward signals.
It can be intractable to obtain effective estimation of value functions
%which can be challenging
in complex problems with highly stochastic dynamics and flawed or delayed reward.
In this section, we propose a way to disentangle the policy-independent part and the reward-independent part during the value estimation process.

Given a trajectory $\tau_{t:t+k} = \left(s_t, a_t, ..., s_{t+k}, a_{t+k} \right)$ with $k \ge 0$,
we consider a representation function $f$ that $m_{t:t+k} = f (\tau_{t:t+k})$, and then introduce the following definitions.
%based on which we define a return function $U$ that gives the corresponding cumulative discounted reward of the trajectory:
%based on which we define a trajectory return function $U$ and a predictive dynamics function $P$ as follows:

%Assume that we have a trajectory return function $U$ and a predictive dynamics function $P$.

\begin{myDefinition}
\label{definition:U}
The trajectory return function $U$ defines the cumulative discounted reward of any trajectory $\tau_{t:t+k}$ with the representation $ m_{t:t+k} = f (\tau_{t:t+k})$:
\begin{equation}
\label{eqation:3}
   U ( m_{t:t+k} ) = r_t + \gamma r_{t+1} + \dots + \gamma^{k} r_{t+k} = \sum_{t^{\prime}=t}^{t+k}\gamma^{t^{\prime} - t} r_{t^{\prime}}.
\end{equation}
\end{myDefinition}
The trajectory return function $U$ models the utility of a trajectory and can be viewed as an imperfect long-term reward model of the environment  since it does not depend on a particular policy.
%In contrast to one-step reward function $\mathcal{R}$, $U$ models the cumulative discounted reward of a trajectory. % instead of the one-step reward.

\begin{myDefinition}
\label{definition:P}
Given the representation function $f$, the predictive dynamics function $P$ denotes the expected representation of the future trajectory for performing action $a \in \mathcal{A}$ in state $s \in \mathcal{S}$, then following a policy $\pi$:
\begin{equation}
\label{eqation:4}
   P^{\pi}( s, a ) = \mathbb{E} \big[ f(\tau_{0:T}) | s_0=s, a_0=a;\pi \big] = \mathbb{E} [ m_{0:T} | s_0=s, a_0=a;\pi ].
\end{equation}
\end{myDefinition}
Note that function $P$ has a similar form with the $Q$-function except for the expectation imposed on the trajectory representation.
It is irrelevant to reward and only predicts how the states and actions would evolve afterwards.
%The predictive dynamics function $P$ is irrelevant with reward and only predicts how the states and actions would evolve afterwards.
%It can be seen as partial dynamics model of the environment which depends on the policy $\pi$.
Now, we can derive the following lemma with the above definitions:

\begin{myLemma}
\label{lemma:lower_bound}
Given a policy $\pi$, the following lower bound of the $Q$-function holds for all $s \in \mathcal{S}$ and $a \in \mathcal{A}$, when function $U$ is convex:
\begin{equation}
    %Q^{\pi}(s,a) = \mathbb{E} \big[U (m_{0:T} )|s_0=s, a_0=a; \pi \big] \ge U \big( \mathbb{E} [ m_{0:T} |s_0=s, a_0=a; \pi ] \big) = U \big(P^{\pi}(s,a) \big).
    Q^{\pi}(s,a) \ge U \big(P^{\pi}(s,a) \big).
\end{equation}
The equality is strictly established when $U$ is a linear function.
\end{myLemma}
The proof can be easily obtained with \emph{Jensen's Inequality}, by replacing the summation in Equation \ref{eqation:1} with function $U$ and then exchanging the expectation and function.
%Complete proof can be found in the Supplementary Material.
Similar conclusion can also be obtained for state value function $V$ and
we focus on the $Q$-function in the rest of the paper.

Lemma \ref{lemma:lower_bound} provides a lower-bound approximation of the $Q$-function as a composite function of $U$ and $P$.
When $U$ is a linear function,
the equality guarantees that we can also obtain the optimal policy through optimizing the composite function.
%a policy that maximizes the lower bound maximizes the true $Q$-function as well.
Since the input of $U$, i.e., $P^{\pi}(s,a)$, can be non-linear,
%can be a non-linear representation,
it still ensures the representation ability of the composite function even with a linear $U$.
%Intuitively, one may consider that the first few layers of a $Q$-function network \cite{Lillicrap2015DDPG,Mnih2015DQN} play a role of function $P$ and the last linear layer works as function $U$.
For the case that $U$ is a commonly adopted ReLU-activated neural network (convex function),
we can still maximize the lower bound of the $Q$-function by maximizing the composite function.
However, there is no guarantee for the optimality of the policy learned in such cases (as we found in our experiments).

The above modeling induces an understanding that the $Q$-function takes an explicit two-step estimation:
1) it first predicts the expected future dynamics under the policy (function $P$),
2) then evaluates the benefits of future prediction (function $U$).
%indicates the understanding of two-step value estimation as mentioned at the beginning of this section.
%Equation 6 indicates an understanding that Q-value function takes a two-step estimation: it first predicts the expected future dynamics (i.e., the representation of state-action trajectories) under the policy and then evaluates the cumulative discounted reward of the future dynamics.
This provides us a way to decompose the value estimation process by dealing with function $P$ and $U$ separately.
%dealing with the above two steps separately.
Thus, prediction and evaluation of state-action trajectory can be more efficient to carry out in a compact representation space.
The decoupling of environmental dynamics and returns helps in stabilizing the value estimation process and provides flexibility for the use in different problems.
Moreover, it draws a connection between model-free RL and model-based RL since the composite function in Lemma \ref{lemma:lower_bound} indicates an evidence of model learning in model-free value estimation.
%We suggest
%that imperfect or partial models of the environment are learned in model-free RL.
Concretely, our decomposition of value functions induces an imperfect reward model $U$ and a partial dynamics model $P$ from the view of trajectory.

Finally, with the composite function approximation in Lemma \ref{lemma:lower_bound},
we can obtain the value-decomposed deterministic policy gradient $\nabla_{\theta} \tilde{J}(\pi_{\theta})$ by extending Equation \ref{eqation:2} accordingly with the Chain Rule:
%the deterministic policy gradient (Equation \ref{eqation:2}) can be extended with the Chain Rule accordingly, with which a policy $\pi_{\theta}$ can be updated:
\begin{equation}
\label{eqation:7}
   \nabla_{\theta} \tilde{J}(\pi_{\theta}) = \mathbb{E}_{s \sim \rho^{\pi}} \left[ \nabla_{\theta} \pi_{\theta}(s)
   \nabla_{a} P^{\pi}(s,a)|_{a=\pi_{\theta}(s)}
   \nabla_{m} U(m)|_{m=P^{\pi}(s,a)}
   \right].
\end{equation}

\section{Algorithm}
In this section, we use the two-step understanding of $Q$-value estimation % understanding of two-step value estimation
discussed in previous section to derive a practical deep RL algorithm based on modern deep learning techniques.

\begin{figure}
\centering
\includegraphics[width=0.95\textwidth]{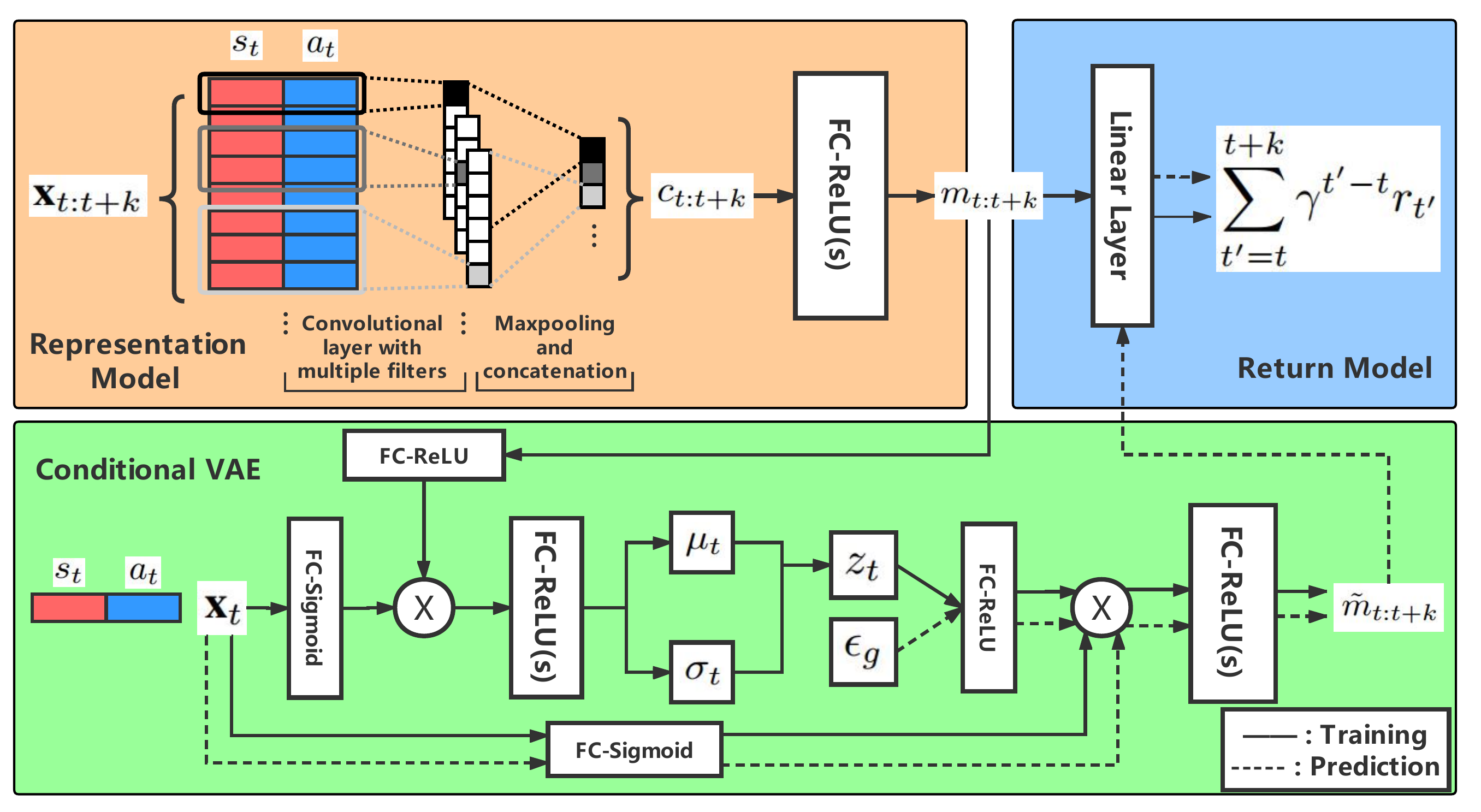}

\caption{The overall network structure of our models: representation model (\emph{orange}), trajectory return model (\emph{blue}) and conditional VAE (\emph{green}).
We abbreviate fully-connected layers as FC (with certain activation) and use $\otimes$ to denote the pairwise-product operation.
The dashed lines illustrate the flow for the two-step prediction of estimated value
with generative decoder ($P$) and return model ($U$).
%with the well-trained models.
}
\label{figure:1}
\end{figure}

\subsection{State-Action Trajectory Representation}
\label{section:representation}
%The trajectory representation function $f$ can be an identical function in the simplest case.
%However,
To derive a practical algorithm, an effective and compact representation function is necessary because: 1) the trajectory may have variant length, 2) and there may be irrelevant features in states and actions which might affect the estimation of the cumulative discounted reward of the trajectory.
In this paper, we propose using Convolutional Neural Networks (CNNs) to learn a representation model $f^{\rm CNN}$ of state-action trajectory, similar to the use in sentence classification \cite{Kim2014CNN}.
In our experiments, we found that this way achieves faster training and better performance than the popular LSTM \cite{Hochreiter1997LSTM} structure (see ablations in Section \ref{section:ablation}).
%Other approaches can also be considered depending on specific problems.
%The representation model is illustrated in the orange part of Figure \ref{figure:1}.
An illustration of $f^{\rm CNN}$ is shown in the orange part of Figure \ref{figure:1}.

Let $\textbf{x}_t \in \mathbb{R}^l$ be the $l$-dimensional feature vector of $(s_t,a_t)$ pair.
A trajectory $\tau_{t:t+k}$ (padded where necessary) is represented as $\textbf{x}_{t:t+k} = \textbf{x}_t \oplus \textbf{x}_{t+1} \oplus \dots \oplus \textbf{x}_{t+k}$, where $\oplus$ is the concatenation operator.
With a convolution filter $\textbf{w}^i \in \mathbb{R}^{h^i \times l}$ applied to a window of $h^i$ state-action pairs and then a max-pooling operation, a feature $c^i_{t:t+k}$ can be generated as follows:
\begin{equation}
\label{eqation:8}
\begin{aligned}
    c^i_j = {\rm ReLU}(\textbf{w}^i \cdot & \textbf{x}_{j:j+h^i-1} + b) \ {\rm where} \ j = t, \dots, t + k - h^i + 1,\\
    & c^i_{t:t+k} = \max \{c^i_t, c^i_{t+1} \dots, c^i_{t + k - h^i + 1}\}.
\end{aligned}
\end{equation}
We apply multiple filters ${\{\textbf{w}^i}\}_{i=1}^{n}$ on $\tau_{t:t+k}$ similarly to generate the $n$-dimensional feature vector
$c_{t:t+k}$,
then obtain the trajectory representation $m_{t:t+k}$ after several fully-connected layers:
\begin{equation}
\label{eqation:9}
\begin{aligned}
    & c_{t:t+k} = (c^1_{t:t+k}, c^2_{t:t+k}, \dots, c^n_{t:t+k}), \\
    m_{t:t+k} = & f^{\rm CNN} (\tau_{t:t+k}) = {\rm FullyConnectedLayers}(c_{t:t+k}).
\end{aligned}
\end{equation}
%In our experiments, we use a max trajectory length and pad the shorter trajectories with all-zero vectors.
%For example, a max length of 256 can be considered to correspond a discounted factor $\gamma = 0.99$ as $0.99^{256} \approx 0.076$.

\subsection{Trajectory Return Model}
\label{section:return}
%In Equation 5 and Equation 6, we use a linear function $U$ for demonstration.
%In Equation \ref{eqation:5} and \ref{eqation:6}, we use a convex function $U$ for demonstration.
Following Lemma \ref{lemma:lower_bound}, we implement the trajectory return function $U$ with convex functions.
Without loss of optimality, we use a linear $U$ to ensure the strict equality, as illustrated in the blue part of Figure \ref{figure:1}.
The result for a popular ReLU-activated layer can be seen in Ablation (Section \ref{section:ablation}).
% and a proof of the lower-bound property can be easily obtained (see more details in appendix) (\textbf{TO-DO}).

We train the representation model $f^{\rm CNN}$ and return model $U^{\rm Linear}$ together by minimizing the mean square error loss of mini-batch samples from experience buffer $\mathcal{D}$, with respect to the parameters $\omega$:
\begin{equation}
\label{eqation:10}
    \mathcal{L}^{\rm Ret}(\omega) = \mathbb{E}_{(\tau_{t:t+k},r_{t:t+k}) \sim \mathcal{D}} \Big[ \big( U^{\rm Linear}(f^{\rm CNN}(\tau_{t:t+k})) - \sum_{t^{\prime}=t}^{t+k}\gamma^{t^{\prime} - t} r_{t^{\prime}} \big)^2 \Big].
\end{equation}

\subsection{Conditional Variational Dynamics Model}
\label{section:conditional-VAE}
The most straightforward way to implement the predictive dynamics function is to use a Multi-Layer Perceptron (MLP) $P^{\rm MLP}$ that takes the state and action as input and predicts the expected representation of future trajectory.
However, such approach does not really model the stochasticity of future trajectory representation.
In this paper, we present a conditional Variational Auto-Encoder (VAE) \cite{Kingma2013AEVB} to model the underlying distribution of future trajectory representation conditioned on the state and action, achieving significant improvement over $P^{\rm MLP}$ (see Ablation in Section \ref{section:ablation}).

The conditional VAE consists of two networks, an encoder $q_{\phi}(z_t|m_{t:t+k}, s_t, a_t)$ and decoder $p_{\varphi}(m_{t:t+k}|z_t, s_t, a_t)$ with variational parameters $\phi$ and generative parameters $\varphi$ respectively.
%, and subscripts are omitted for clarity.
With a chosen prior, generally the multivariate normal distribution $\mathcal{N}(0,I)$, the encoder approximates the conditional posteriors of latent variable $z_t$ for trajectory representation, producing a Gaussian distribution with mean $\mu_t$ and standard deviation $\sigma_t$.
The decoder generates a representation of future trajectory $\tilde{m}_{t:t+k}$ for a given latent variable conditioned on the state-action pair.
Besides, we use a pairwise-product operation to emphasize an explicit relation between the condition stream and trajectory representation stream, which shows better inference results in our experiments (see Ablation in Section \ref{section:ablation}).
The structure of conditional VAE is illustrated in the green part of Figure \ref{figure:1}.

During training,
%the latent variable is sampled from $\mathcal{N}(\mu_t, \sigma_t)$ with reparameterization trick \cite{Kingma2013AEVB}, i.e., $z_t = \mu_t + \sigma_t \cdot \mathcal{N}(0, I)$.
the latent variable is sampled from $\mathcal{N}(\mu_t, \sigma_t)$ with reparameterization trick \cite{Kingma2013AEVB}, i.e., $z_t = \mu_t + \sigma_t \cdot \mathcal{N}(0, I)$, which is taken as part of input by the decoder to reconstruct the trajectory representation.
This naturally models the underlying stochasticity of future trajectory.
We train the conditional VAE with respect to the variational lower bound \cite{Kingma2013AEVB}, in a form of the reconstruction loss along with a KL divergence term (see the Supplementary Material for complete formulation):
\begin{equation}
\label{eqation:11}
    \mathcal{L}^{\rm VAE}(\phi, \varphi) = \mathbb{E}_{\tau_{t:t+k} \sim \mathcal{D}} \Big[ \|m_{t:t+k} - \tilde{m}_{t:t+k}\|_2^2
    + \beta D_{\rm KL} \big( \mathcal{N}(\mu_t,\sigma_t) \| \mathcal{N}(0,I) \big) \Big],
\end{equation}
where $m_{t:t+k}$ is obtained from the representation model (Equation \ref{eqation:9}).
We use a weight $\beta > 1$ to encourage VAE to discover disentangled latent factors for better inference quality, which is also known as a $\beta$-VAE \cite{Burgess2018Understanding,Higgins2017Beta}.
See Ablation (Section \ref{section:ablation}) for the results of different values of $\beta$.

%For generation process, a noise sampled from $\mathcal{N}(0,I)$ is passed into the decoder.
Since VAE infers the latent distribution via instance-to-instance reconstruction, during the generation process,
we propose using a clipped generative noise to narrow down the discrepancy between the generated instance and the expected representation (Equation \ref{eqation:4}).
This allows us to obtain high-quality prediction of expected future trajectory.
Finally, the predictive dynamics model $P^{\rm VAE}$ can be viewed as the generation process of the conditional VAE with a clipped generative noise $\epsilon_g$:
%:
%the decoder takes a state-action pair $(s,a)$ as the condition and a generative Gaussian noise sampled from $\mathcal{N}(0,I)$ to predict the expected representation of future trajectory.
%given a state-action pair $(s,a)$ as the condition and a clipped generation noise $\epsilon_g$:
\begin{equation}
\label{eqation:12}
    \tilde{m}_{t:t+k} = P^{\rm VAE}(s, a, \epsilon_{g}), {\rm and} \ \epsilon_{g} \sim {\rm Clip} \big( \mathcal{N}(0,I), -c, c \big).
\end{equation}
%Since VAE infers the distribution of future trajectory representation, we clip the noise during generation process with $c$ to ensure the generated (predicted) representation close to the expected representation rather than a representation instance.
When $c$ is zero, an expected representation of future trajectory (Equation \ref{eqation:4}) should be generated from the mean of the latent distribution.
A further discussion of clip value $c$ is in Ablation (Section \ref{section:ablation}).

\subsection{Overall Algorithm}
We build our algorithm on Deep Deterministic Policy Gradient (DDPG) \cite{Lillicrap2015DDPG} algorithm, by replacing the original critic (i.e., the $Q$-function) with a decomposed one, consisting of the three models introduced in previous subsections.
%The Q-value function is trained in a two-fold way and the actor is updated similarly with respect to the decomposed critic.
%The three models are trained from the experiences collected during interaction and
The actor is updated through gradient ascent (Equation \ref{eqation:7}) similarly with respect to the decomposed critic.
Note our algorithm does not use target networks for both the actor and critic. % since no TD error \cite{Sutton1988ReinforcementLA} needs to be calculated here as in DDPG.
%by following the extended deterministic policy gradient in Equation 7.
The overall algorithm is summarized in Algorithm \ref{algorithm:vdfp}.
\begin{algorithm}
  \caption{Value decomposed DDPG with future prediction (VDFP) algorithm}
  \begin{algorithmic}[1]
    \State Initialize actor network $\pi_\theta$ with random parameters $\theta$, and experience buffer $D$
    \State Initialize representation model and trajectory return model with random parameters $\omega$
    \State Initialize conditional VAE with random parameters $\phi$, $\varphi$
    %\State Initialize experience buffer $D$
	\For{episode $= 1, 2, \dots$}
	    \For{$t = 1, 2, \dots, T$}
	        \State Observe state $s_t$ and select action $a_t = \pi_{\theta}(s_t) + \epsilon_{\rm e}$, with exploration noise $\epsilon_{\rm e} \sim \mathcal{N}(0,\sigma)$
	        \State Execute action $a_t$ and obtain reward $r_t$
	        %\State Sample mini-batch of $N$ experience $\{ ( \tau_{t_i:t_i+k} ) \}_{i=1}^{N}$ from $\mathcal{D}$
	        \State Sample mini-batch of $N$ experience $\{ ( \tau_{t_i:T} ) \}_{i=1}^{N}$ from $\mathcal{D}$
	        \State Update conditional VAE by minimizing $\mathcal{L}^{\rm VAE}(\phi, \varphi)$
	        \State Predict the representation of future trajectory $\tilde{m} = P^{\rm VAE}(s,a,\epsilon_{g})|_{s=s_i,a=\pi_{\theta}(s_i)}$
	        \State Update actor $\pi_{\theta}$ with the value-decomposed deterministic policy gradient (Equation \ref{eqation:7})%, with predicted representation $\tilde{m} = P^{\rm VAE}(s,a,\epsilon_{g})|_{s=s_i,a=\pi_{\theta}(s_i)}$, $\epsilon_{g} \sim {\rm Clip} \big( \mathcal{N}(0,1), -0.2, 0.2 \big)$
        \EndFor
        %\State Store experiences $\{ (\tau_{t:t+k}, r_{t:t+k}) \}_{t=1}^{T}$ in $\mathcal{D}$
        \State Store experiences $\{ (\tau_{t:T}, r_{t:T}) \}_{t=1}^{T}$ in $\mathcal{D}$
        \For{epoch $= 1, 2, \dots$, num\_epoch}
            %\State Sample mini-batch of $N$ experience $\{ (\tau_{t_i:t_i+k}, r_{t_i:t_i+k}) \}_{i=1}^{N}$ from $\mathcal{D}$
            \State Sample mini-batch of $N$ experience $\{ (\tau_{t_i:T}, r_{t_i:T}) \}_{i=1}^{N}$ from $\mathcal{D}$
    	    \State Update representation model $f^{\rm CNN}$ and return model $U^{\rm Linear}$ by minimizing $\mathcal{L}^{\rm Ret}(\omega)$
        \EndFor
	\EndFor
  \end{algorithmic}
\label{algorithm:vdfp}
\end{algorithm}

\section{Related Work}

\paragraph{Future Prediction}
Thinking about the future has been considered as an integral component of human cognition \cite{Atance2001EpisodicFT,Schacter2007TheCN}.
In neuroscience, one concept named the prospective brain
\cite{Schacter2007RememberingTP} indicates that a crucial function of the brain is to use stored information to predict possible future events.
The idea of future prediction is also studied in model-based RL \cite{Atkeson1997Robot,Sutton1991Dyna}.
%In model-based RL, the idea of future prediction is also studied to learn the dynamics model of the environment for synthetic experience generation or planning \cite{Atkeson1997Robot,Sutton1991Dyna}.
Simulated Policy Learning \cite{Kaiser2019MBRL} is proposed to learn one-step predictive models from the real environment and then train a policy within the simulated environment.
Multi-steps and long-term future are also modeled in \cite{Hafner2019Learning,Ke2019LearningD} with recurrent variational dynamics models,
%Hafner et. al. \cite{Hafner2019Learning} propose the Deep Planning Network (PlaNet) that learns a Recurrent State Space Model to predict multi-step environment dynamics given a sequence of actions.
after which actions are chosen through online planning with Model-Predictive Control (MPC).
%Similarly with a recurrent variational dynamics model and MPC planning, Ke et. al. \cite{Ke2019LearningD} models the long-term future with the augmentation of backward recurrent state.
%In this paper, we look inside model-free RL and provide an explanation of value function estimation from the future prediction perspective.
Besides, another related work is \cite{Dosovitskiy2017DFP}, in which a supervised model is trained to predict the residuals of goal-related measurements at a set of temporal offsets in the future.
With a manually designed goal vector, actions are chosen to maximize the predicted outcomes.
%We consider this approach as a special form of

\paragraph{Value Function Approximation}
Most model-free deep RL algorithms approximate value functions directly with deep neural networks,
e.g., Proximal Policy Optimization (PPO) \cite{Schulman2017PPO}, Advantage Actor-Critic (A2C) \cite{Mnih2016AC} and DDPG \cite{Lillicrap2015DDPG},
without explicitly handling the coupling of environmental dynamics and rewards.
%, which are popular especially in continuous control tasks.
One similar approach to our work is the Deep Successor Representation (DSR) \cite{Kulkarni2016DSR}, which factors the value function into the dot-product between the expected representation of state occupancy and a vector of immediate reward function.
The representation is trained with TD algorithm \cite{Sutton1988ReinforcementLA} and the vector is approximated from one-step transitions.
In our work, we decompose the value function based on the composite form of trajectory dynamics and returns.
In contrast to using TD algorithm, we use a conditional VAE to model the latent distribution and obtain expected trajectory representation.
%The return model, along with the representation of trajectories are trained from trajectory samples.
%Value function decomposition is also studied in multiagent RL to investigate the relation between individual value functions of cooperative agents and the joint value function \cite{Rashid2018QMix,Sunehag2018VDN}.
We demonstrate the superiority of our algorithm in the experimental section.

%Our work is also related to model-free deep RL approaches especially in continuous control tasks, e.g.,
% have achieved a lot of success .

\section{Experiments}

We conduct our experiments on MuJoCo continuous control tasks in OpenAI gym \cite{Brockman2016Gym,Todorov2012MuJoCo}.
%MuJoCo is a commonly adopted environment in RL studies of continuous control tasks and a convenient comparison can be obtain by referring to \cite{Fujimoto2018TD3,Lillicrap2015DDPG,Schulman2017PPO}.
For the convenience of reproducibility, we make no modifications to the original environments or reward functions (except the delay reward modification in Section \ref{delay-reward-section}).
Open source code and learning curves are provided in the Supplementary Material and will soon be released on GitHub.

\subsection{Evaluation}
\label{section:evaluation}

To evaluate the effectiveness of our algorithm,
we focus on two representative MuJoCo tasks: HalfCheetah-v1 and Walker2d-v1, as adopted in \cite{Fujimoto2018TD3,Fujimoto2018BCQ,Schulman2015TRPO,Schulman2017PPO}.
%Each task is run for 1 million time steps and the results are reported over 5 random seeds of Gym simulator and network initialization.
We compare our algorithm (VDFP) against DDPG, PPO, A2C, as well as the Deterministic DSR (DDSR).
For PPO and A2C, we adopt non-parallelization implementation and use Generalized Advantage Estimation \cite{Schulman2016GAE} with $\lambda = 0.95$ for stable policy gradient.
Since DSR is originally proposed based on DQN \cite{Mnih2015DQN} for discrete action problems, we implement DDSR based on DDPG according to the author's codes for DSR on GitHub.
For VDFP, we set the KL weight $\beta$ as 1000 and the clip value $c$ as 0.2.
We use the max trajectory length of 64 and 256 for HalfCheetah-v1 and Walker2d-v1 respectively.
For VDFP, DDPG and DDSR, a Gaussian noise sampled from $\mathcal{N}(0,0.1)$ \cite{Fujimoto2018TD3} is added to each action for exploration.
For all algorithms, we use a two-layer feed-forward neural network of 200 and 100 hidden units with ReLU activation for both the actor and critic (similar scales for the critic variants in DDSR and VDFP).
%as we found Ornstein-Uhlenbeck noise \cite{Uhlenbeck1930OU} which originally used in DDPG offered no apparent benefits.
%For VDFP, we set the KL weight $\beta$ as 1000 and the clip value $c$ as 0.2,
%and we consider a max trajectory length and use zero-padding when necessary.
%For example, a max length of 256 can be considered to correspond a discounted factor $\gamma = 0.99$, as $0.99^{256} \approx 0.076$.

\begin{figure}
\centering
\subfigure[HalfCheetah-v1]{
\includegraphics[width=0.47\textwidth]{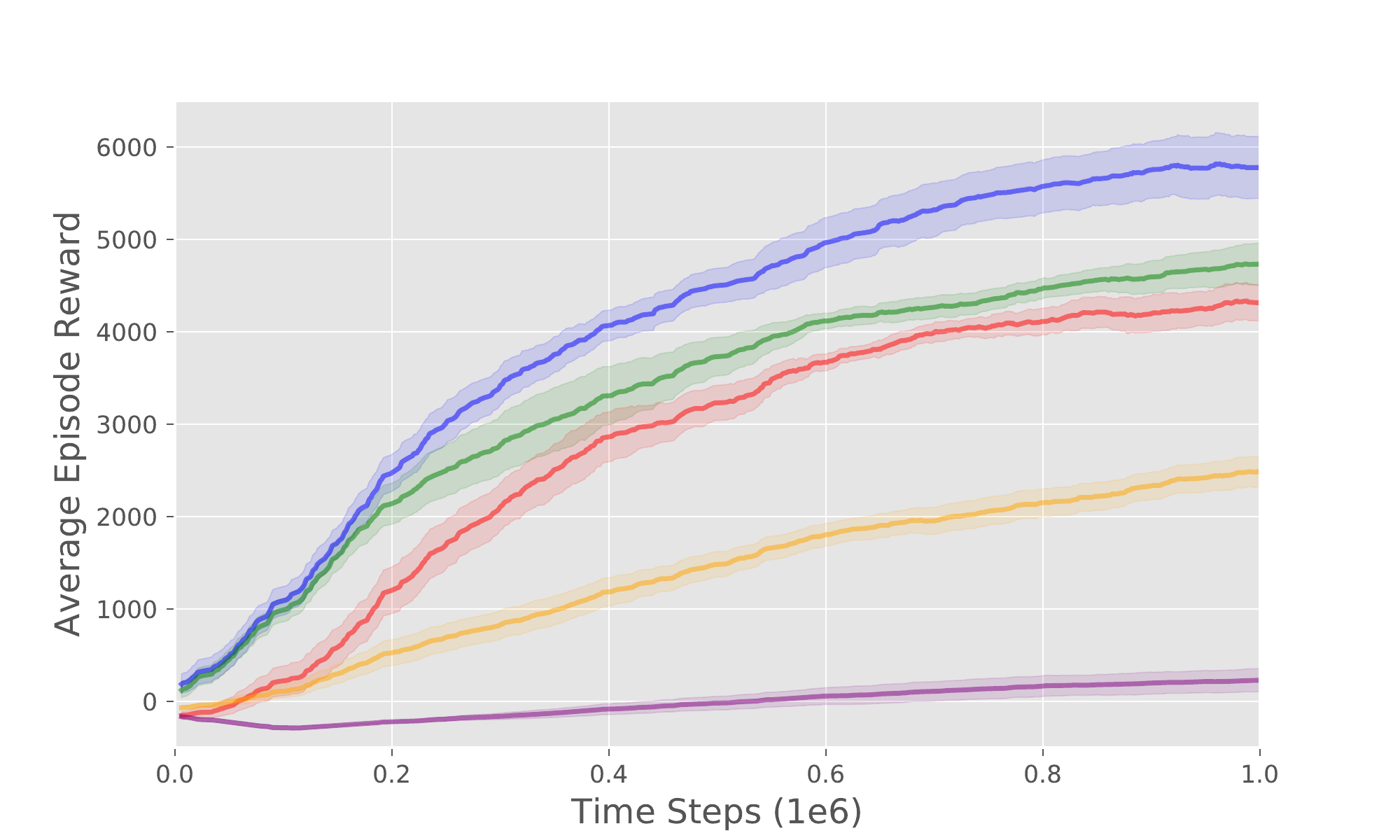}
}
\hspace{-1.0cm}
\subfigure[Walker2d-v1]{
\includegraphics[width=0.53\textwidth]{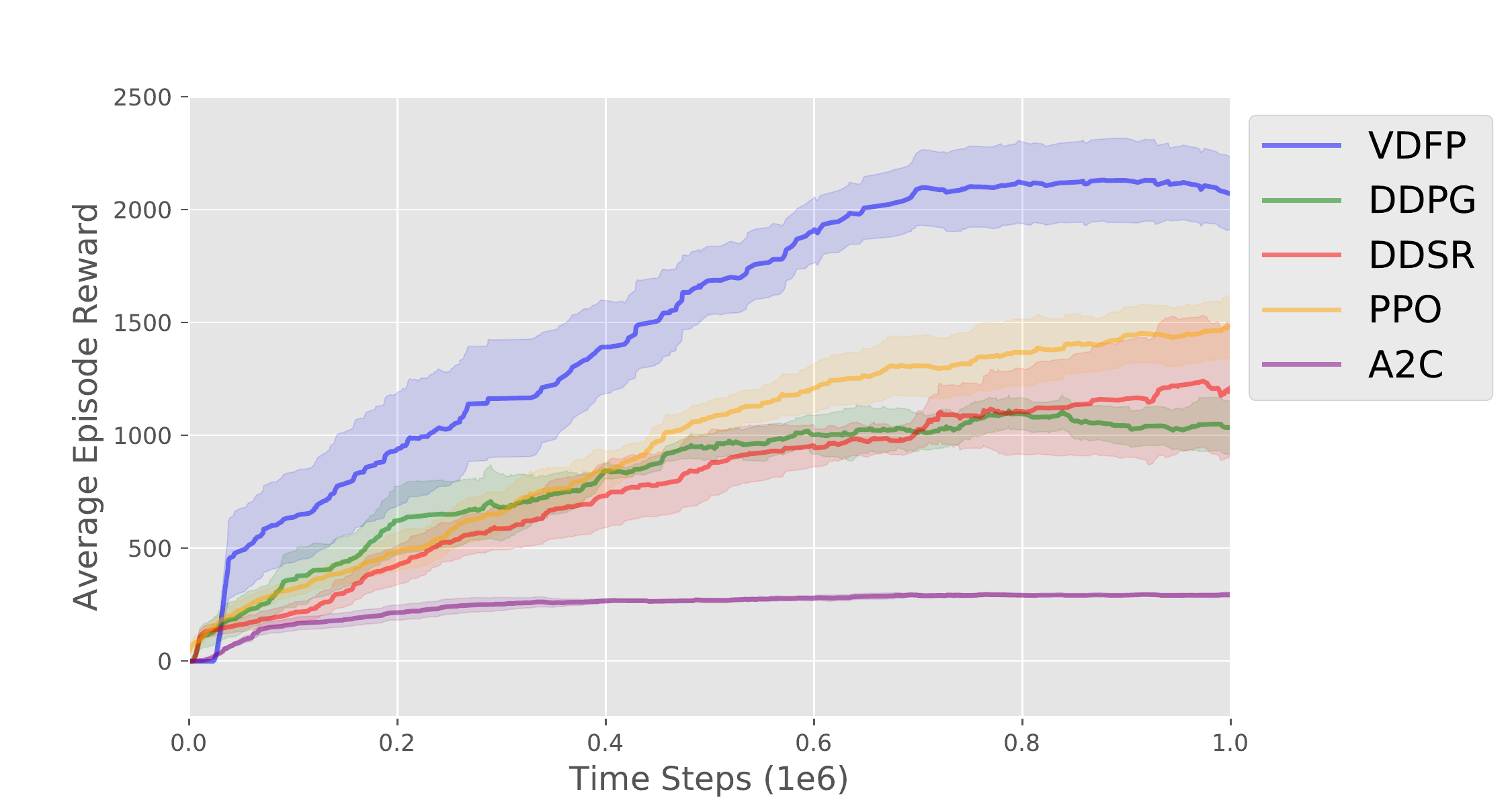}
}

\caption{Learning curves of algorithms in MuJoCo tasks.
The shaded region denotes half a standard deviation of average evaluation over 5 trials.
Results are smoothed over recent 100 episodes.}
%\caption{Cooperative games.}
\label{figure:2}
\end{figure}

%Each task is run for 1 million time steps and
%the results are reported over 5 random seeds of Gym simulator and network initialization.
Figure \ref{figure:2} shows learning curves of algorithms over 5 random seeds of the Gym simulator and the network initialization.
%For VDFP, we use the max trajectory length of 64 and 256 for HalfCheetah-v1 and Walker2d-v1 respectively.
%Each task is run for 1 million time steps and the results are reported over 5 random seeds of the Gym simulator and the network initialization.
We can observe that our algorithm (VDFP) outperforms other algorithms in both final performance and learning speed.
Our results for DDPG, PPO and A2C are comparable with those in \cite{Fujimoto2018TD3,Schulman2017PPO}, where other results for ACKTR \cite{Wu2017ACKTR} and TRPO \cite{Schulman2015TRPO} can also been found.
Exact experimental details of each algorithm are provided in the Supplementary Material.

\subsection{Ablation}
\label{section:ablation}

We perform ablation studies to analyze the contribution of each component of VDFP:
1) CNN (v.s. LSTM) for trajectory representation model (Representation);
2) conditional VAE (v.s. MLP) for predictive dynamics model (Architecture);
3) pairwise-product (v.s. concatenation) operation for conditional encoding process (Operator);
and 4) a single linear layer (v.s. ReLU-activated layers) for trajectory return model (Return).
We use the same experimental setups in Section \ref{section:evaluation} and the results are presented in Table \ref{table:ablation}.
Complete learning curves can be found in the Supplementary Material.

First, we can observe that CNN achieves better performance than LSTM.
%indicating that effective representation can be learned.
Actually, CNN also shows lower training losses and takes less practical training time (almost 8x faster) in our experiments.
%This means that CNN is able to learn effective features of state-action trajectories.
Second, the significance of conditional VAE is demonstrated by its superior performance over MLP.
This is because it is difficult for MLP to approximate the expected representation from various trajectory instances.
In contrast, conditional VAE can well capture the trajectory distribution and then obtain the expected representation through the generation process.
Third, pairwise-product shows an obvious improvement over concatenation.
We suggest that the explicit relation between the condition and representation imposed by pairwise-product, forces the conditional VAE to learn more effective hidden features.
Lastly, adopting linear layer for the trajectory return model outperforms the case of using ReLU-activated layers since it ensures the equality between the composite function approximation and the $Q$-function (Lemma \ref{lemma:lower_bound}), thus obtains a better guarantee for the optimal policy.

Moreover, we analyse the influence of weight $\beta$ for KL loss term (Equation \ref{eqation:11}) and clip value $c$ for prediction process (Equation \ref{eqation:12}).
The results for different values of $\beta$ are consistent to the studies about $\beta$-VAE \cite{Burgess2018Understanding,Higgins2017Beta}:
larger $\beta$ applies stronger emphasis on VAE to discover disentangled latent factors, resulting in better inference performance.
For clip value $c$, clipping achieves superior performance than not clipping ($c = \infty$) since this narrows down the discrepancy between prediction instance and expected representation of future trajectory as we discussed in Section \ref{section:conditional-VAE}.
Though the complete clipping ($c = 0.0$) should ensure the consistence to the expected representation and shows a good performance and lower deviation,
%Though complete clipping should ensure the consistence to the expected representation,
considering the imperfect approximation of neural networks,
setting $c$ to a small positive value ($c = 0.2$) actually achieves a slightly better result.

\begin{table}
  \caption{Ablation of VDFP across each contribution in HalfCheetah-v1.
  Results are Max Average Episode Reward over 5 trials of 1 million time steps.
  $\pm$ corresponds to half a standard deviation.
  Note that Operator, KL weight and clip value are not applicable (N/A) for the MLP architecture here.
  }
  \label{table:ablation}
  \centering
  \scalebox{0.8}{
  \begin{tabular}{cc|cc|cc|cc|cc|c}
    \toprule
    \multicolumn{2}{c}{Representation} &  \multicolumn{2}{c}{Architecture} & \multicolumn{2}{c}{Operator} & \multicolumn{2}{c}{Return}                  \\
    \cmidrule(r){1-2} \cmidrule(r){3-4} \cmidrule(r){5-6} \cmidrule(r){7-8}
    CNN     & LSTM & VAE & MLP & Pairwise-Prod. & Concat. & Linear & ReLU & $\beta$ & $c$ & Results\\
    \midrule
    \checkmark & & \checkmark & & \checkmark & & \checkmark & & 1000 & 0.2 & \textbf{5818.60} $\pm$ \ \textbf{336.25}\\
    \midrule
        & \checkmark & \checkmark & & \checkmark & & \checkmark & & 1000 & 0.2 & 5197.03 $\pm$ \ 156.52\\
    \checkmark & & & \checkmark & N/A & N/A & \checkmark &  & N/A & N/A & 2029.00 $\pm$ \ 486.11\\
    \checkmark & & \checkmark & & & \checkmark & \checkmark & & 1000 & 0.2 & 4541.71 $\pm$ \ 104.22\\
    \checkmark & & \checkmark & & \checkmark & & & \checkmark & 1000 & 0.2 & 5119.04 $\pm$ \ 390.89\\
    \midrule
    \checkmark & & \checkmark & & \checkmark & & \checkmark & & 100 & 0.2 & 4794.96 $\pm$ \ 370.02\\
    \checkmark & & \checkmark & & \checkmark & & \checkmark & & 10 & 0.2 & 3933.33 $\pm$ \ 361.82\\
    \midrule
    \checkmark & & \checkmark & & \checkmark & & \checkmark & & 1000 & $\infty$ & 4752.84 $\pm$ 328.75\\
    \checkmark & & \checkmark & & \checkmark & & \checkmark & & 1000 & 0.0 & 5712.55 $\pm$ \ 233.74\\
    \bottomrule
  \end{tabular}
  }
\end{table}

\subsection{Delayed Reward}
\label{delay-reward-section}

%In previous two subsections, we evaluate VDFP under common settings.
%we demonstrate that VDFP achieves superior performance under the common setting.
We further demonstrate the significant effectiveness and robustness of VDFP under delayed reward settings.
We consider two representative delayed reward settings in real-world scenarios:
1) multi-step accumulated rewards are given at sparse time steps;
2) each one-step reward is delayed for certain time steps.
To simulate above two settings, we make a simple modification to MuJoCo tasks respectively:
1) deliver $d$-step accumulated reward every $d$ time steps and at the end of an episode;
2) delay the immediate reward of each step by $d$ steps and compensate at the end of episode.

\begin{figure}
\centering
\hspace{-0.0cm}
\subfigure[delay step $d$ = 16]{
\begin{minipage}{0.245\textwidth}
\includegraphics[width=1\textwidth]{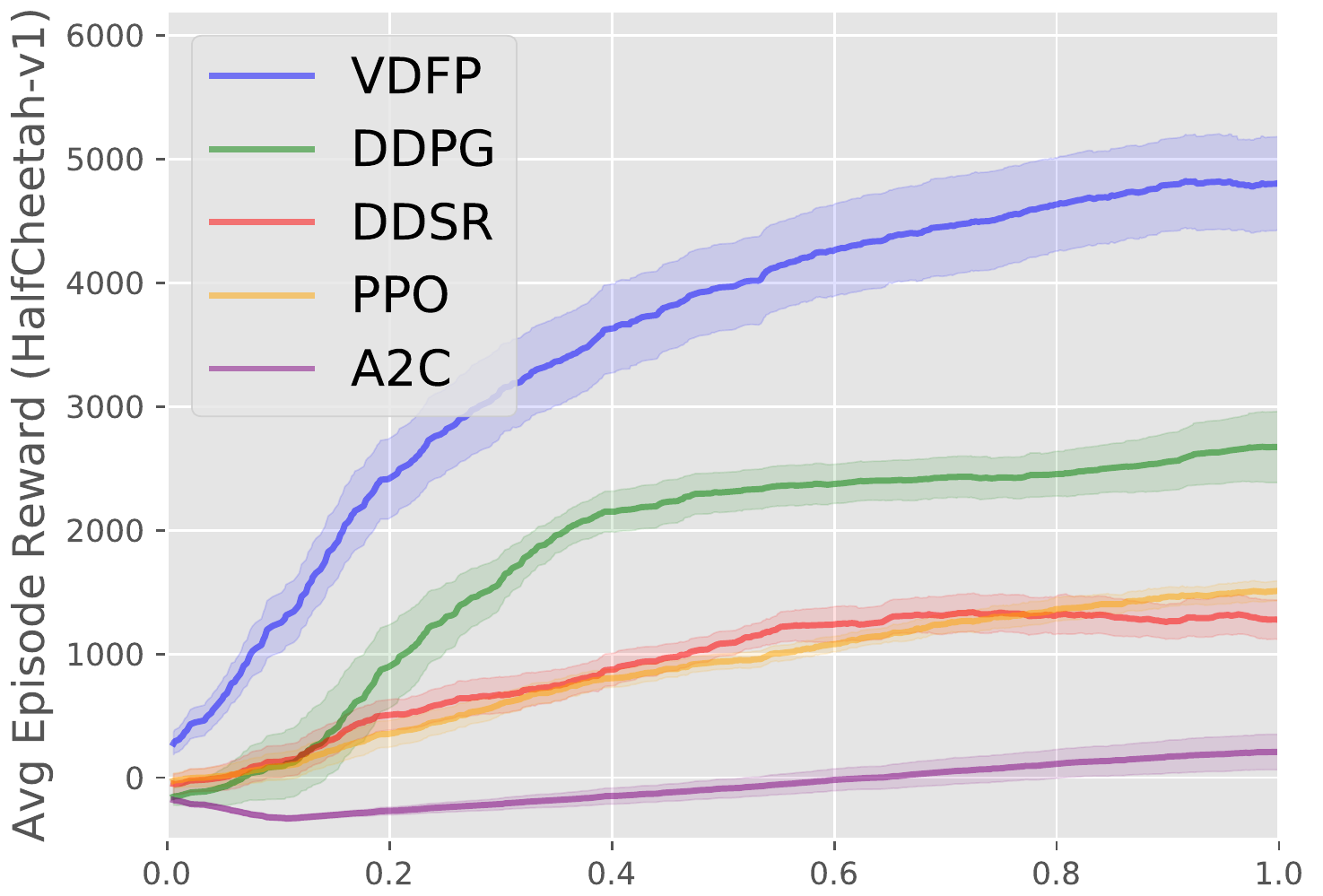} \\
\includegraphics[width=1\textwidth]{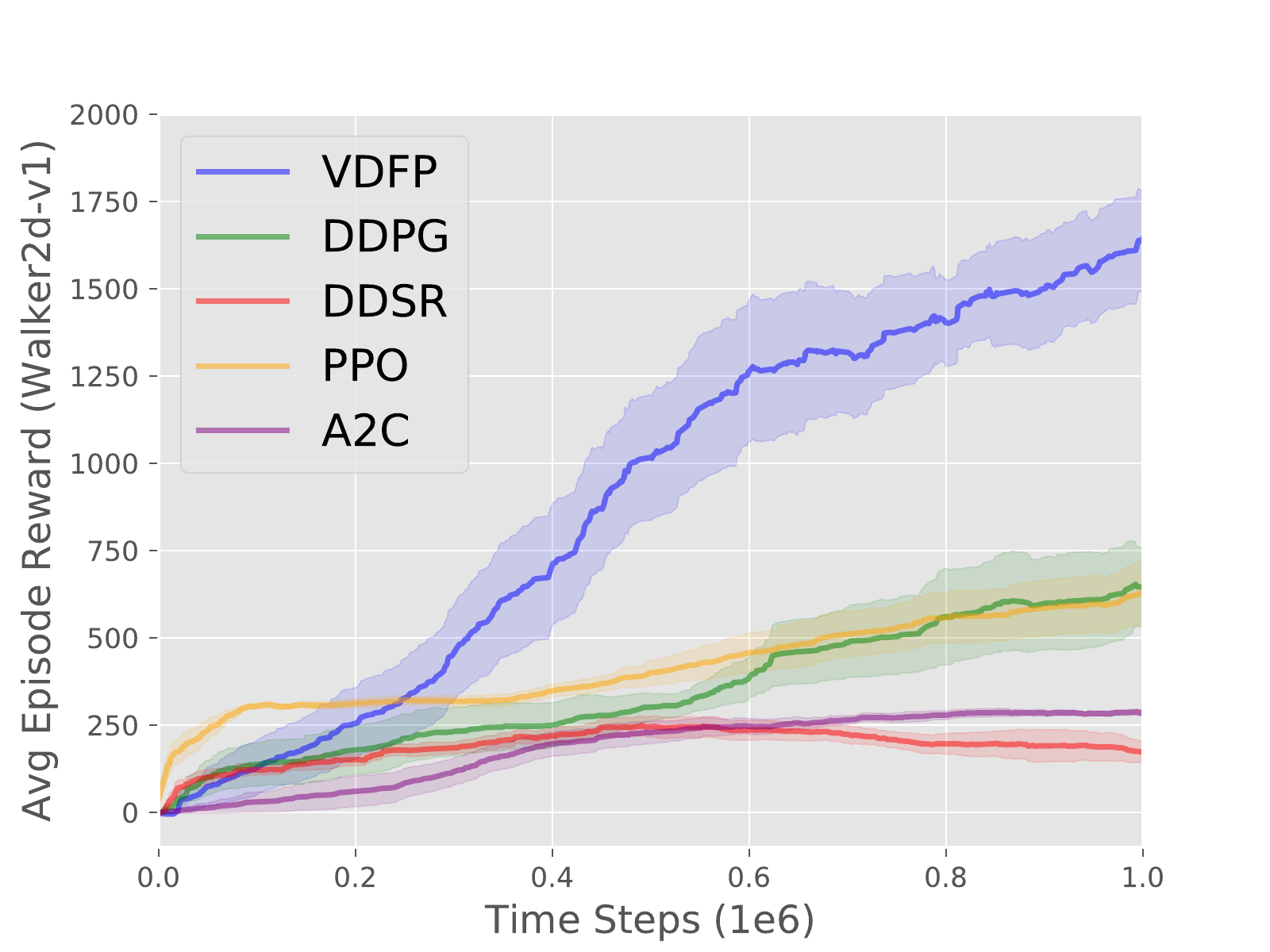}
\end{minipage}
}
\hspace{-0.3cm}
\subfigure[delay step $d$ = 32]{
\begin{minipage}{0.235\textwidth}
\includegraphics[width=1\textwidth]{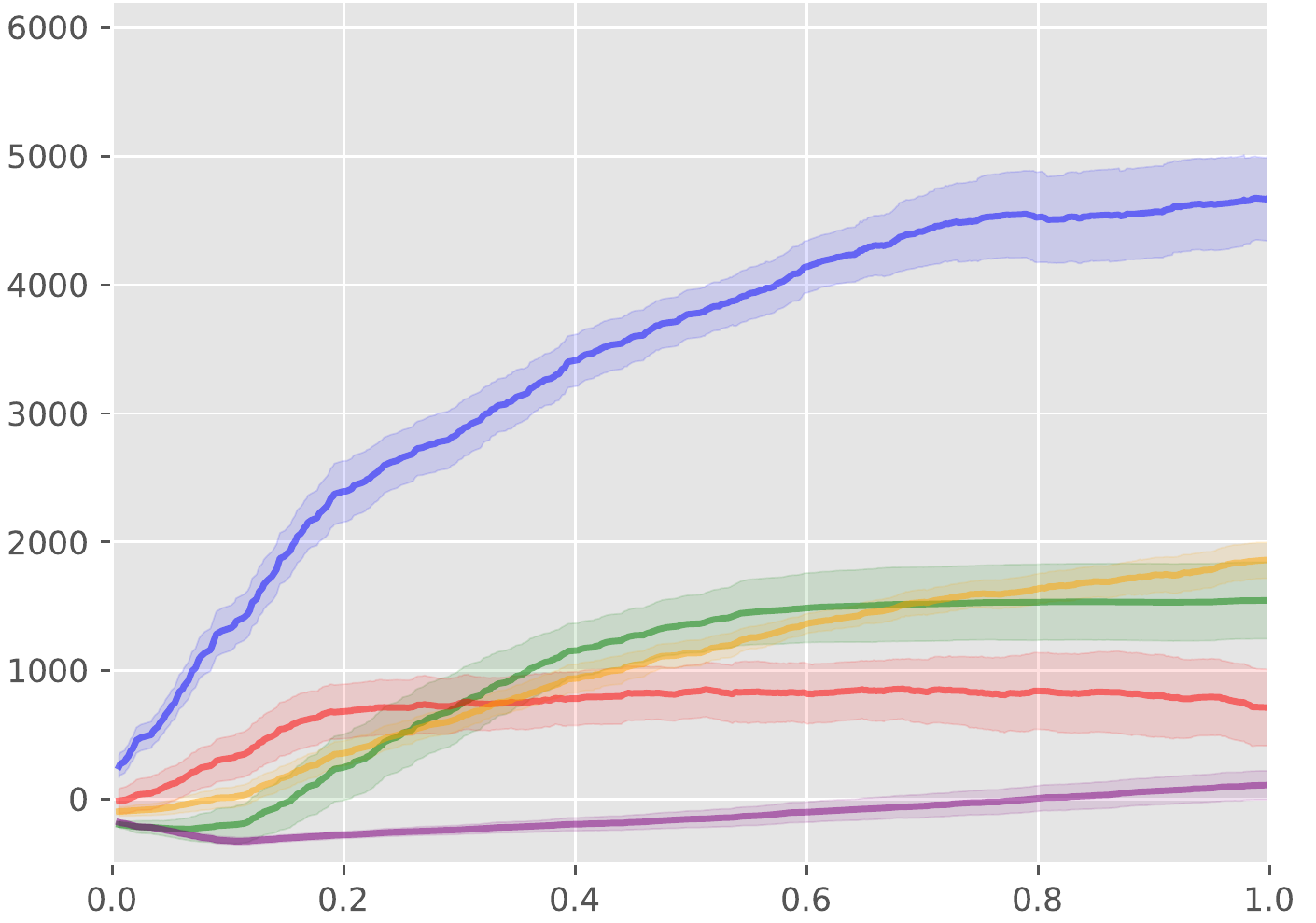} \\
\includegraphics[width=1\textwidth]{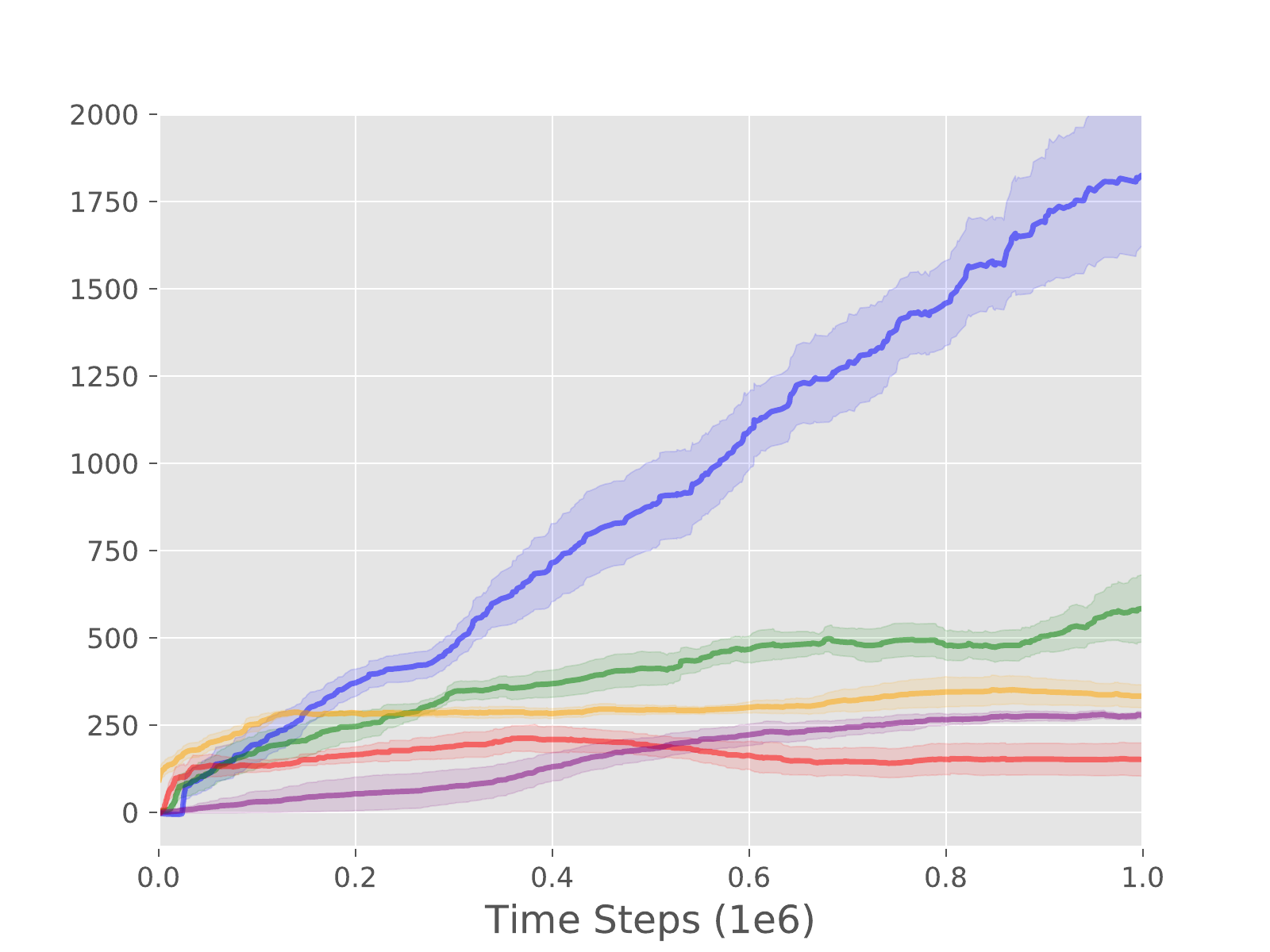}
\end{minipage}
}
\hspace{-0.3cm}
\subfigure[delay step $d$ = 64]{
\begin{minipage}{0.235\textwidth}
\includegraphics[width=1\textwidth]{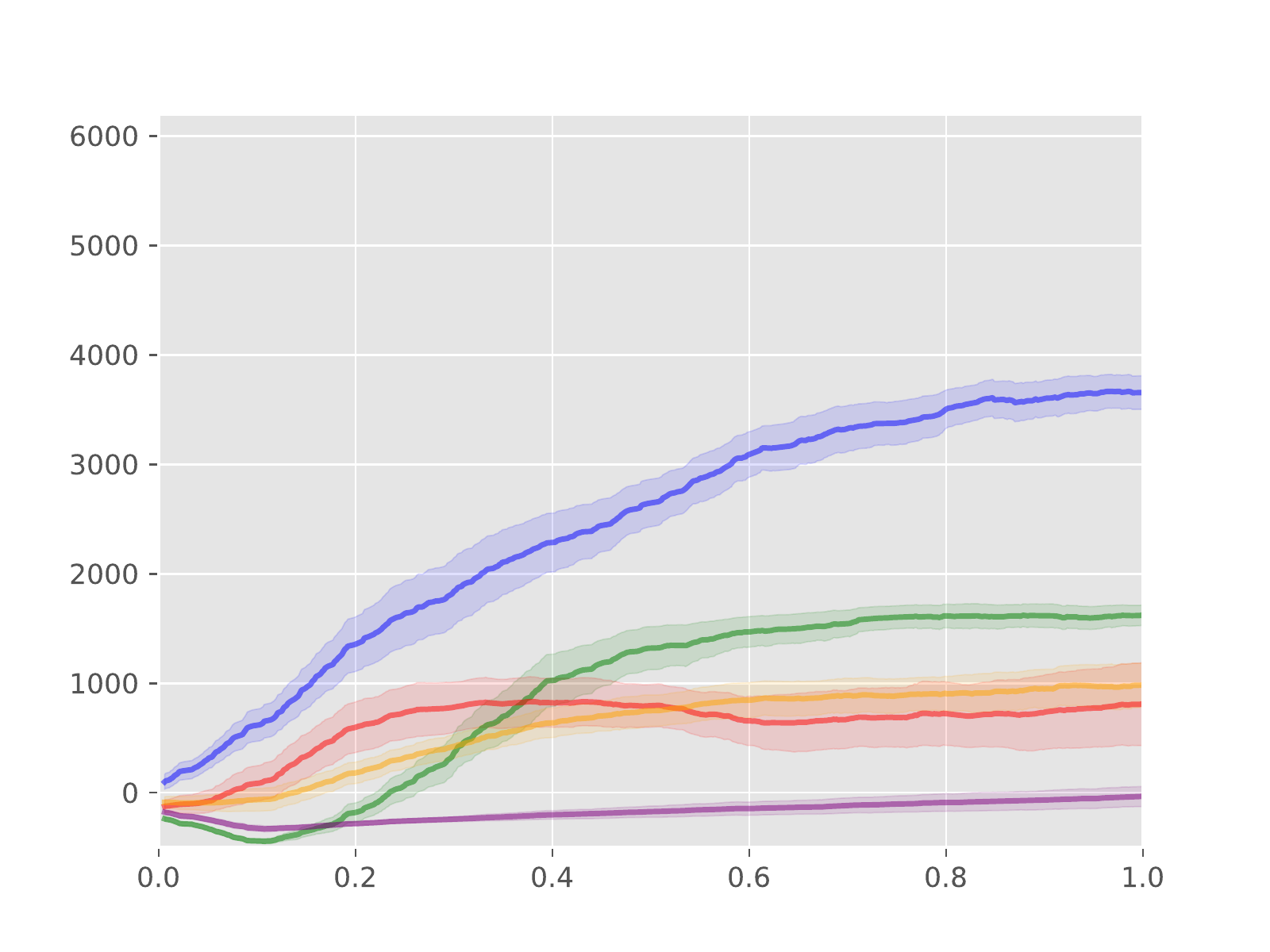} \\
\includegraphics[width=1\textwidth]{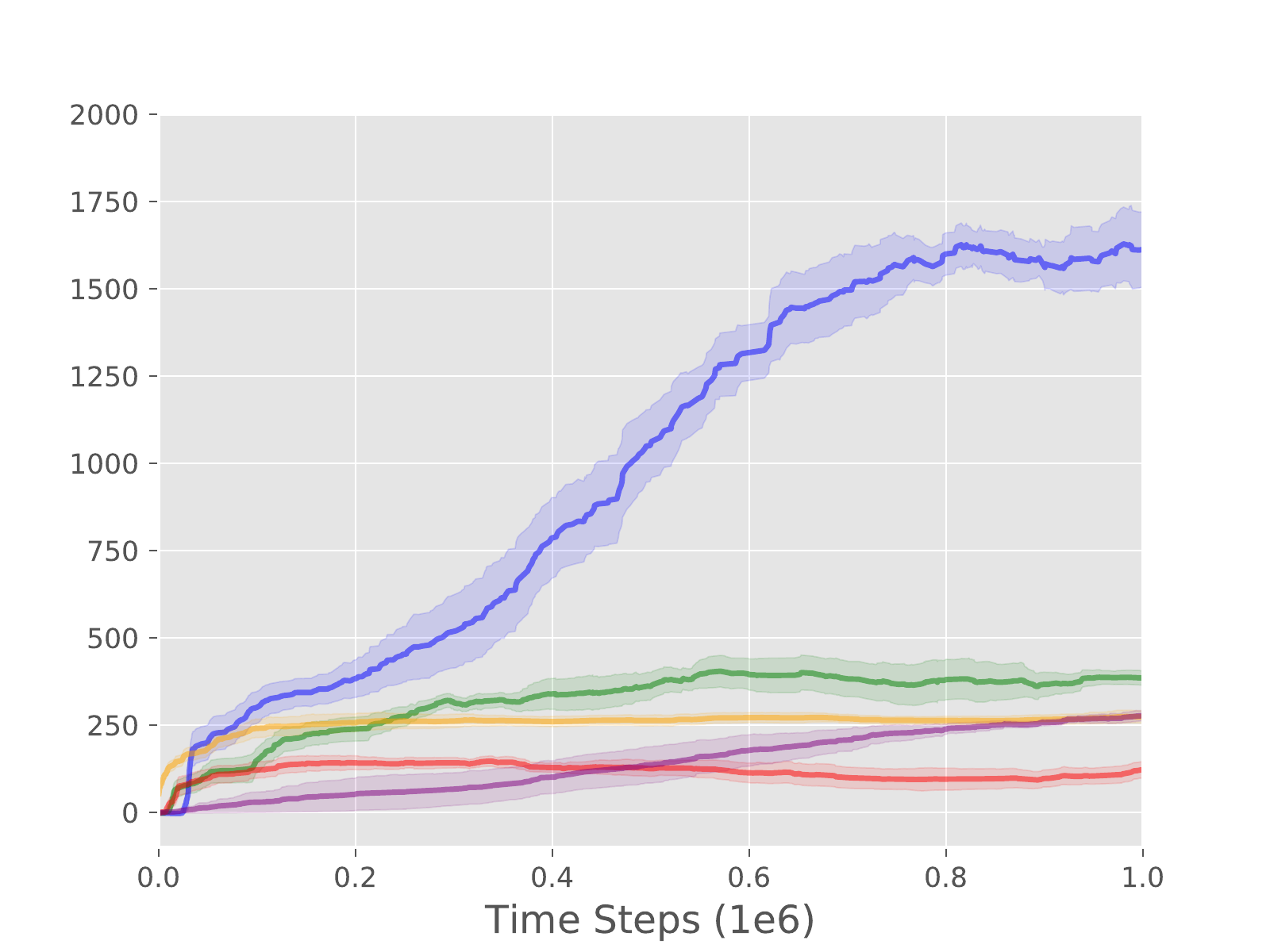}
\end{minipage}
}
\hspace{-0.3cm}
\subfigure[delay step $d$ = 128]{
\begin{minipage}{0.235\textwidth}
\includegraphics[width=1\textwidth]{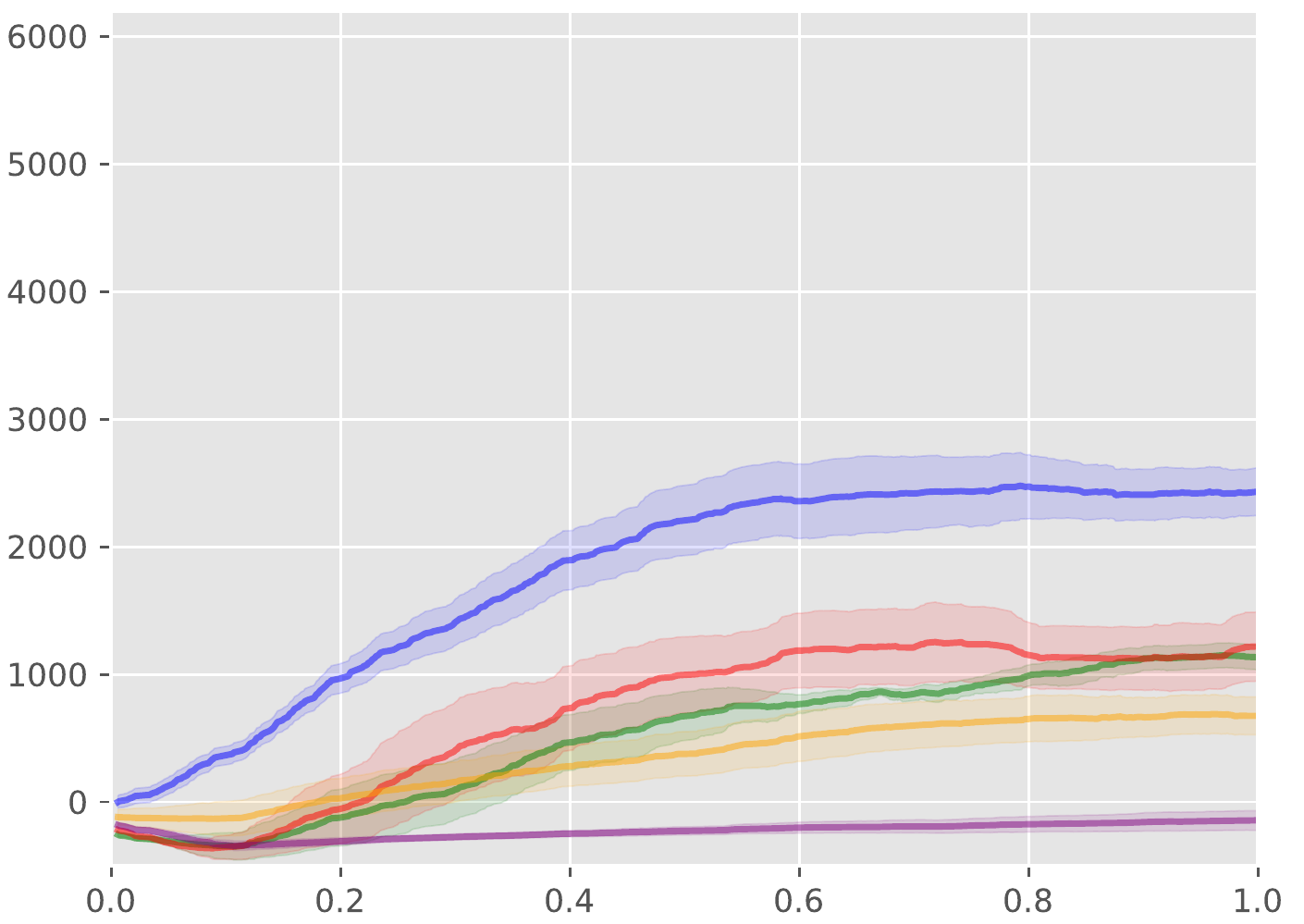} \\
\includegraphics[width=1\textwidth]{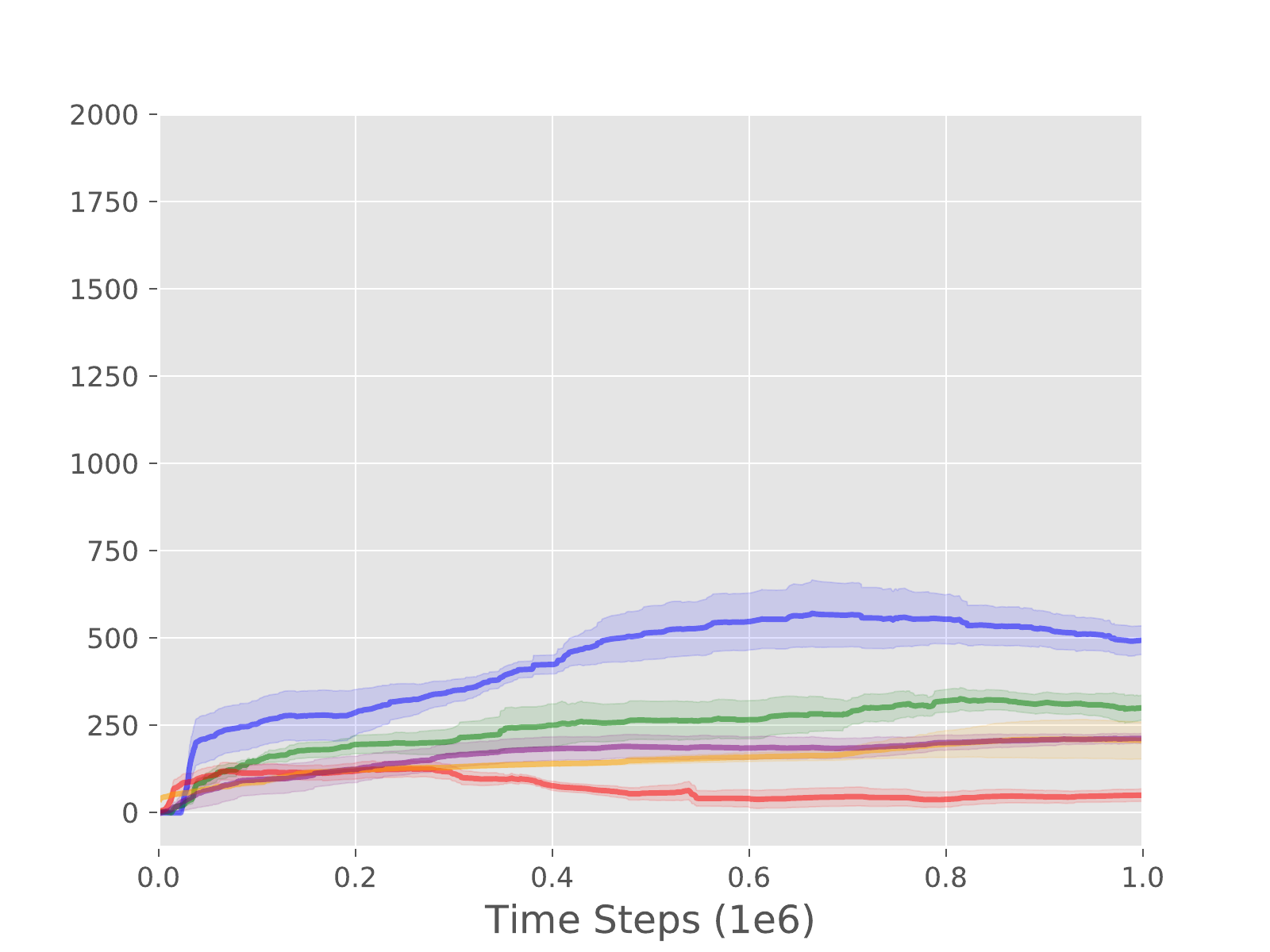}
\end{minipage}
}

\caption{
Learning curves under the first delayed reward setting.
Different delay steps are listed from left to right.
The upper and lower rows are for HalfCheetah-v1 and Walker2d-v1 respectively.
}
\label{figure:3}
\end{figure}

With the same experimental setups in Section \ref{section:evaluation}, we evaluate the algorithms under different delayed reward settings, with a delay step $d$ from 16 to 128.
For VDFP, a max trajectory length of 256 is used for all settings
except that using 64 for HalfCheetah-v1 with $d =$ 16 and 32 already ensures a good performance.
Figure \ref{figure:3} plots the results under the first delayed reward setting.
Similar results are also observed for the second class of delay reward and can be found in the Supplementary Material.
%Figure \ref{figure:3} plots the results for the two delayed reward settings in HalfCheetah-v1.
% and are omitted due to space limitation.
%Similar results for Walker2d-v1 and complete curves can be found in the Supplementary Material.

As the increase of delay step $d$, all algorithms gradually degenerate in comparison with Figure \ref{figure:2} ($d = 0$).
DDSR can hardly learn effective policies under such delayed reward settings due to the failure of its one-step reward model even with a relatively small delay step (e.g., $d$ = 16).
VDFP consistently outperforms others under all settings, in both learning speed and final performance (2x to 4x than DDPG).
Besides, VDFP shows good robustness with delay step $d \le 64$.
As discussed in Section \ref{section:model}, we suggest that the reason for the superior performance of VDFP is two-fold:
1) VDFP can always learn the dynamics of the environment effectively from state and action feedbacks, which is irrelevant with how rewards are delayed actually;
2) the trajectory return model is robust with delayed reward since it approximates the cumulative reward instead of one-step immediate reward.

\section{Conclusion and Future Work}

We present an explicit two-step understanding of value functions in model-free RL from the perspective of future prediction.
Through re-writing the value function in a composite function form, we decompose the value estimation process into two separate parts, which allows more effective and flexible use in different problems.
Further, we derive our algorithm from such decomposition and innovatively propose a conditional variational dynamics model with clipped generation noise to predict the future.
Evaluation and ablation studies are conducted in MuJoCo continuous control tasks.
The effectiveness and robustness are also demonstrated under challenging delay reward settings.

In this paper, we use a off-policy training for VDFP and it could be flawed since trajectories collected by old policies may not able to represent the future under current policy.
However, we do not observe any adverse effect of using off-policy training in our experiments (similar results are also found in \cite{Dosovitskiy2017DFP}), and explicitly introducing several on-policy correction approaches shows no apparent benefits.
We hypothesize that it is because the deterministic policy used in VDFP relaxes the on-policy requirements.
%We remain the further investigation of this issue and the extension to stochastic policy as future work.
It is worthwhile further investigation of this issue and the extension to stochastic policy.
Besides, to some extend, the variational predictive dynamics model of VDFP can be viewed as a Monte Carlo (MC) based estimation \cite{Sutton1988ReinforcementLA} over the space of trajectory representation.
In traditional RL approaches, MC value estimation are widely known to suffer from high variance.
Thus, we suggest that VDFP may indicate a new variational MC approach with lower variance.
We consider the theoretical analysis of the variance reduction as another future work.

\small
\bibliographystyle{plain}
\bibliography{NIPS19-VDFP}

\clearpage
\appendix

\section*{A. Complete Learning Curves}

\subsection*{A.1. Learning Curves for the Results in Ablation}
\begin{figure}[h]
\centering
\subfigure[VAE v.s. MLP]{
\includegraphics[width=0.34\textwidth]{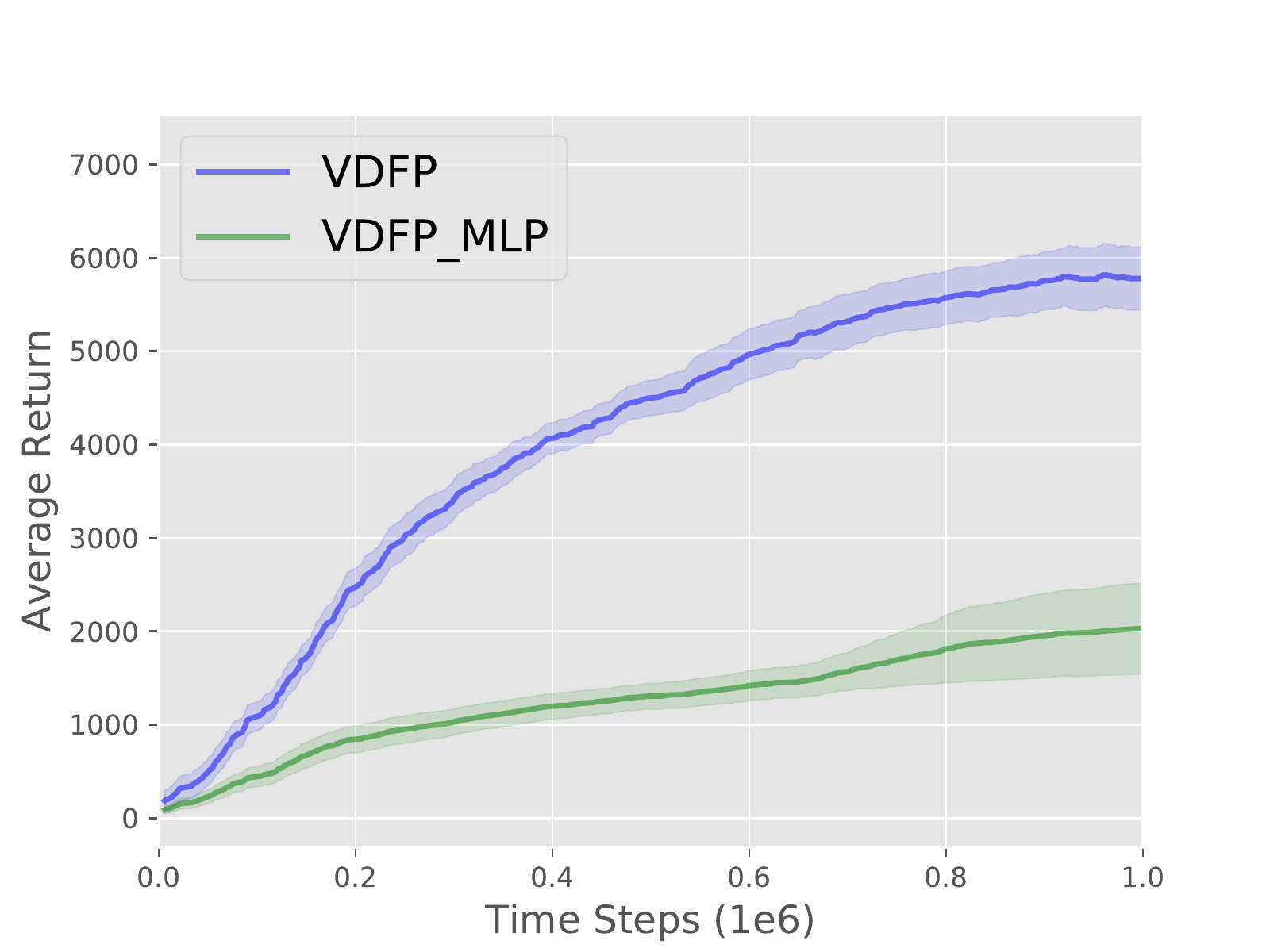}
}
\hspace{-0.8cm}
\subfigure[CNN v.s. LSTM]{
\includegraphics[width=0.34\textwidth]{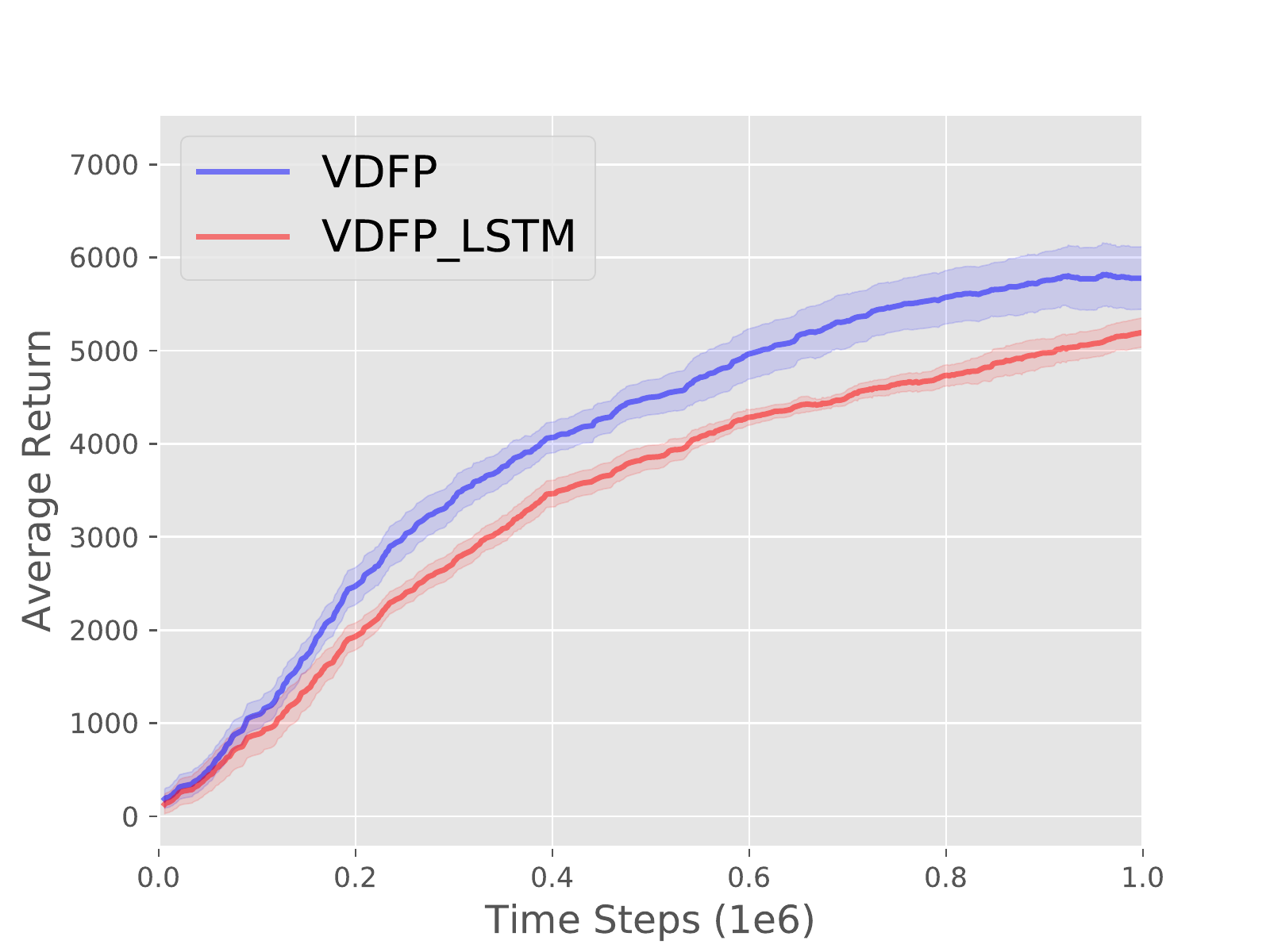}
}
\hspace{-0.8cm}
\subfigure[Pairwise-Prod. v.s. Concat.]{
\includegraphics[width=0.34\textwidth]{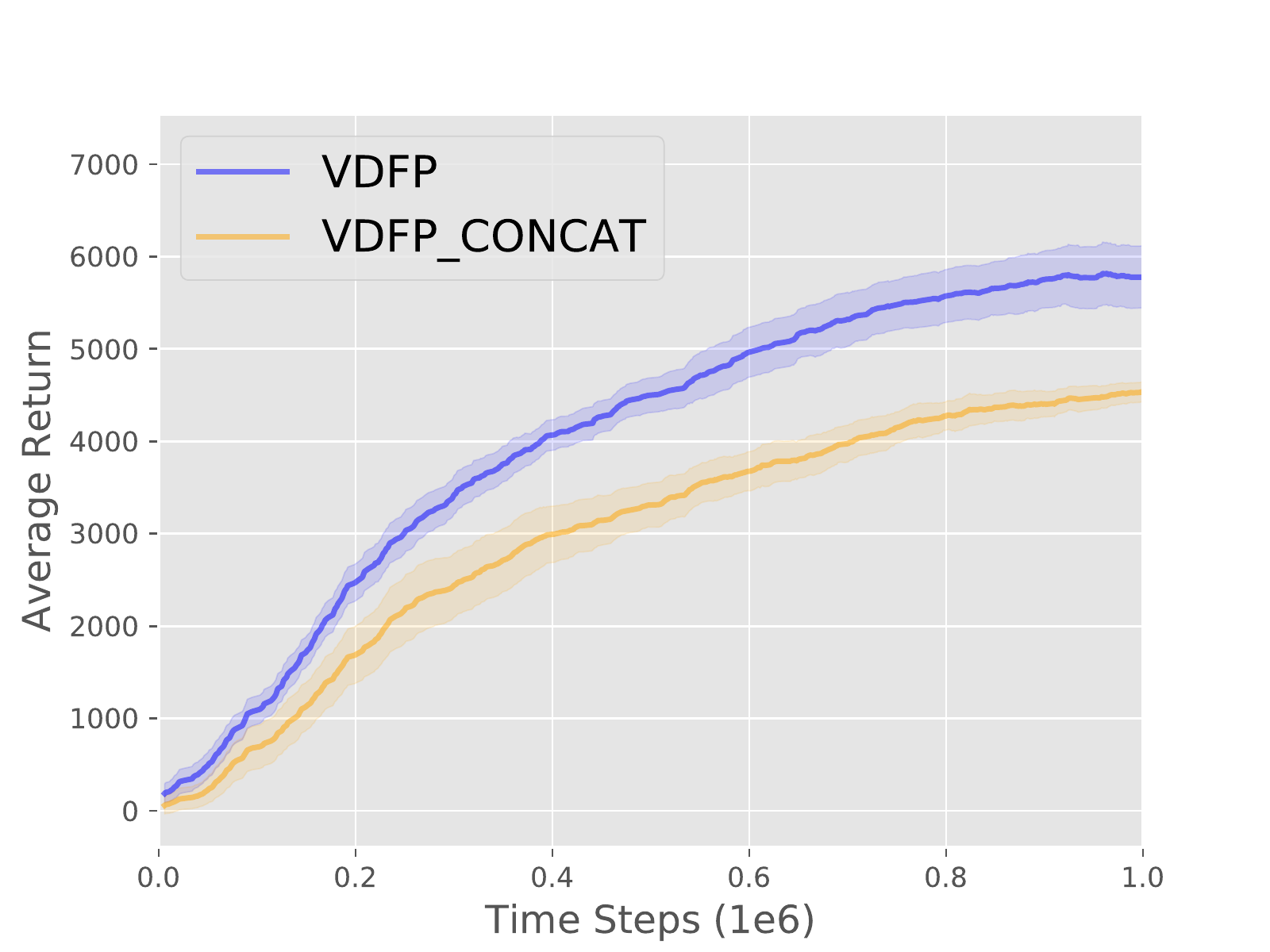}
}
\hspace{-0.8cm}
\subfigure[Linear v.s. ReLU]{
\includegraphics[width=0.34\textwidth]{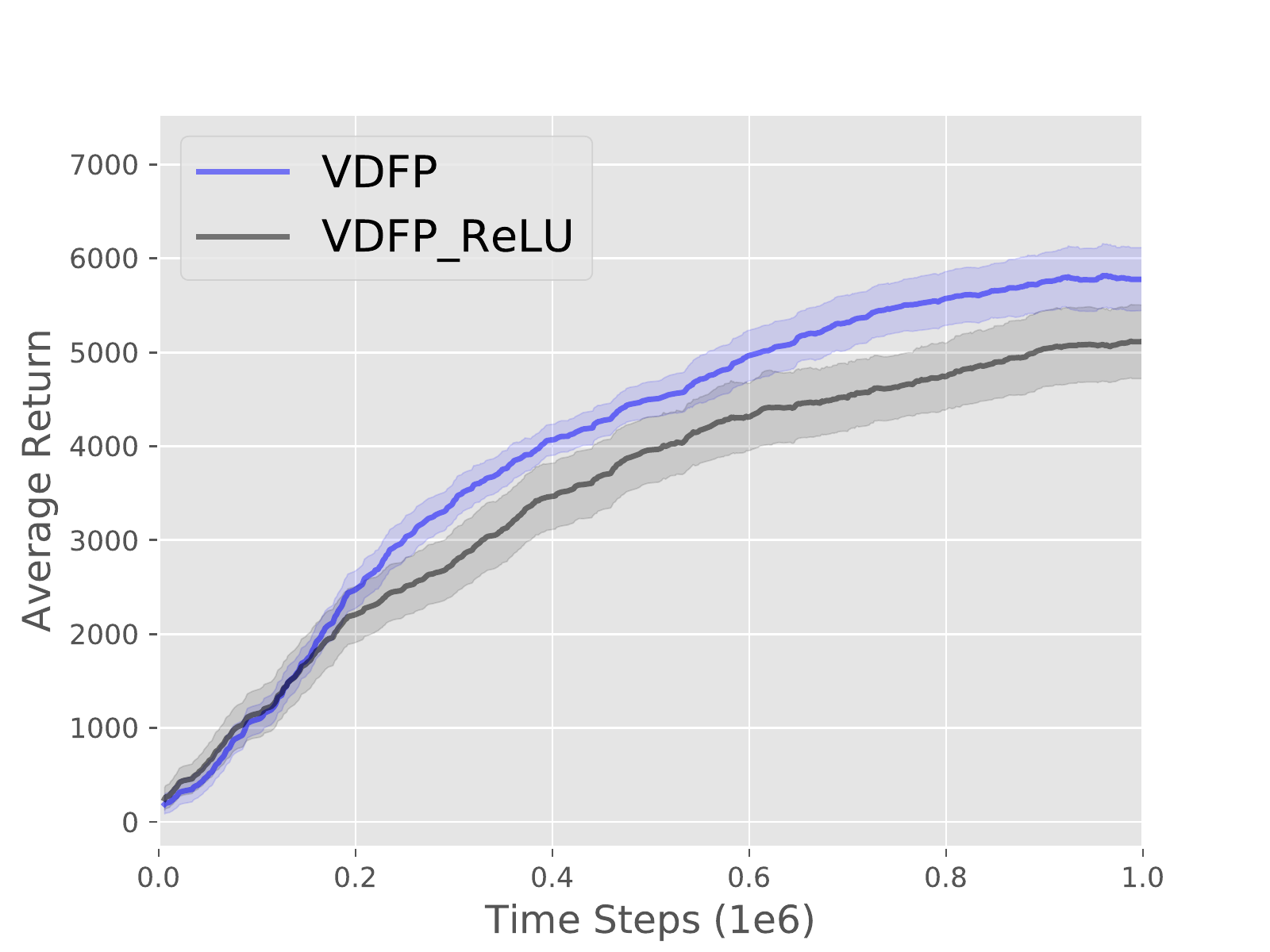}
}
\hspace{-0.8cm}
\subfigure[KL weights]{
\includegraphics[width=0.34\textwidth]{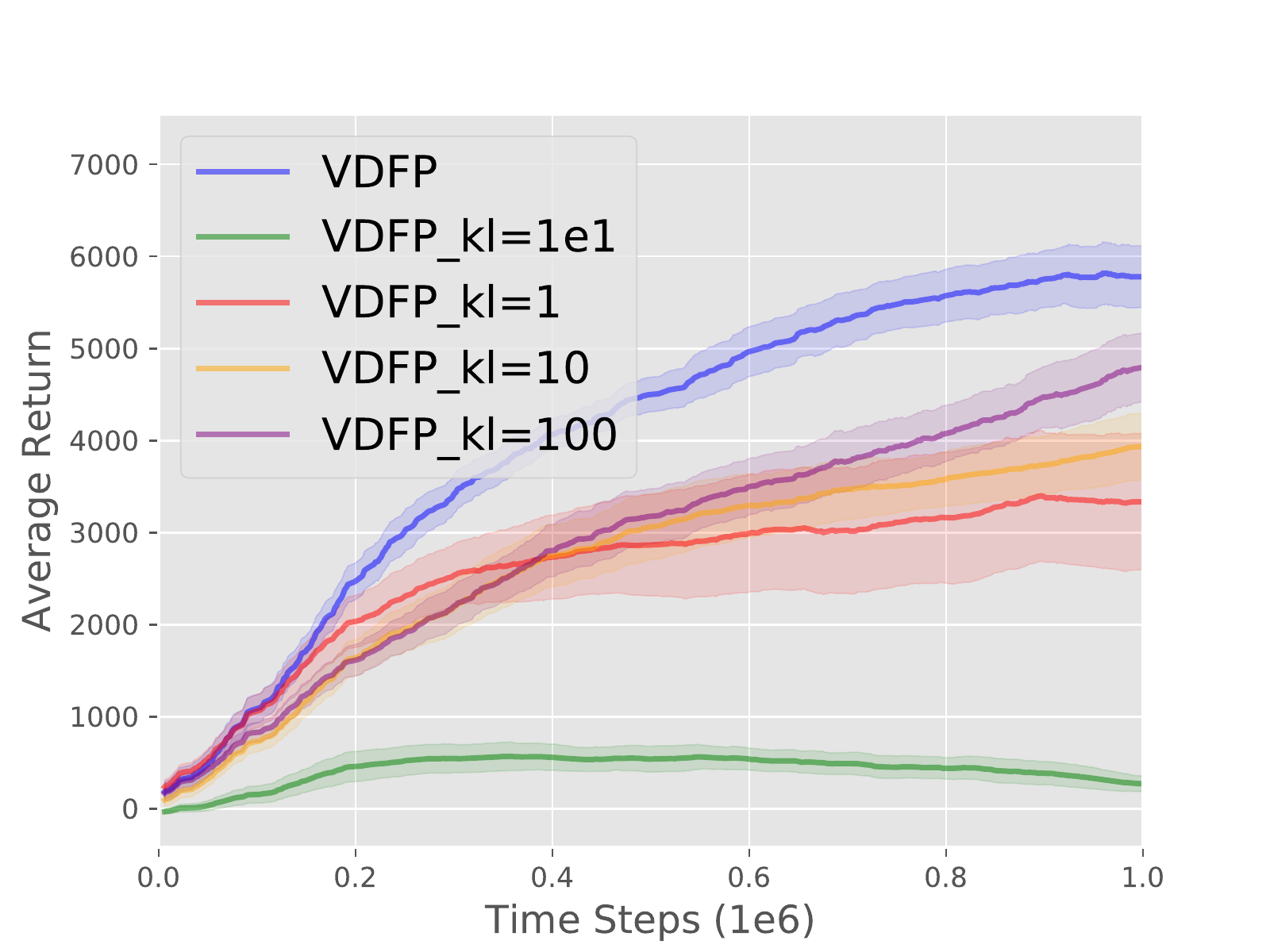}
}
\hspace{-0.8cm}
\subfigure[Clip value]{
\includegraphics[width=0.34\textwidth]{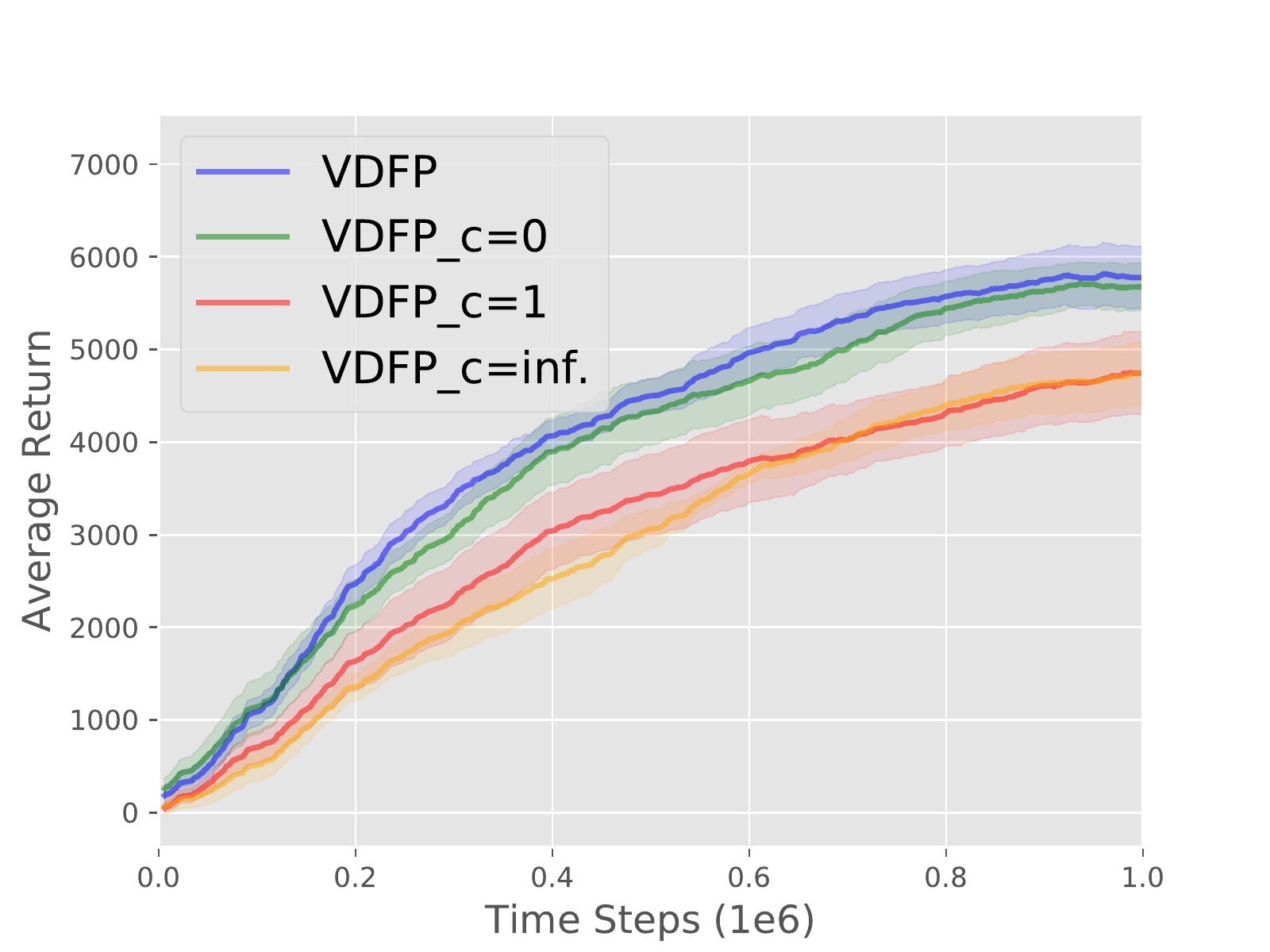}
}

\caption{Learning curves of ablation studies for VDFP (i.e., VAE + CNN + Pairwise-Product + linear layer + kl=1000 + c=0.2) in HalfCheetah-v1.
The shaded region denotes half a standard deviation of average evaluation over 5 trials.
Results are smoothed over recent 100 episodes.}
%\caption{Cooperative games.}
\label{app_figure:ablation}
\end{figure}

Figure \ref{app_figure:ablation} shows the learning curves of VDFP and its variants for ablation studies (Section \ref{section:ablation}), corresponding to the results in Table \ref{table:ablation}.

\subsection*{A.2. Results for the Second Delayed Reward Setting}
\begin{figure}
\centering
\subfigure[delay step $d$ = 0]{
\includegraphics[width=0.33\textwidth]{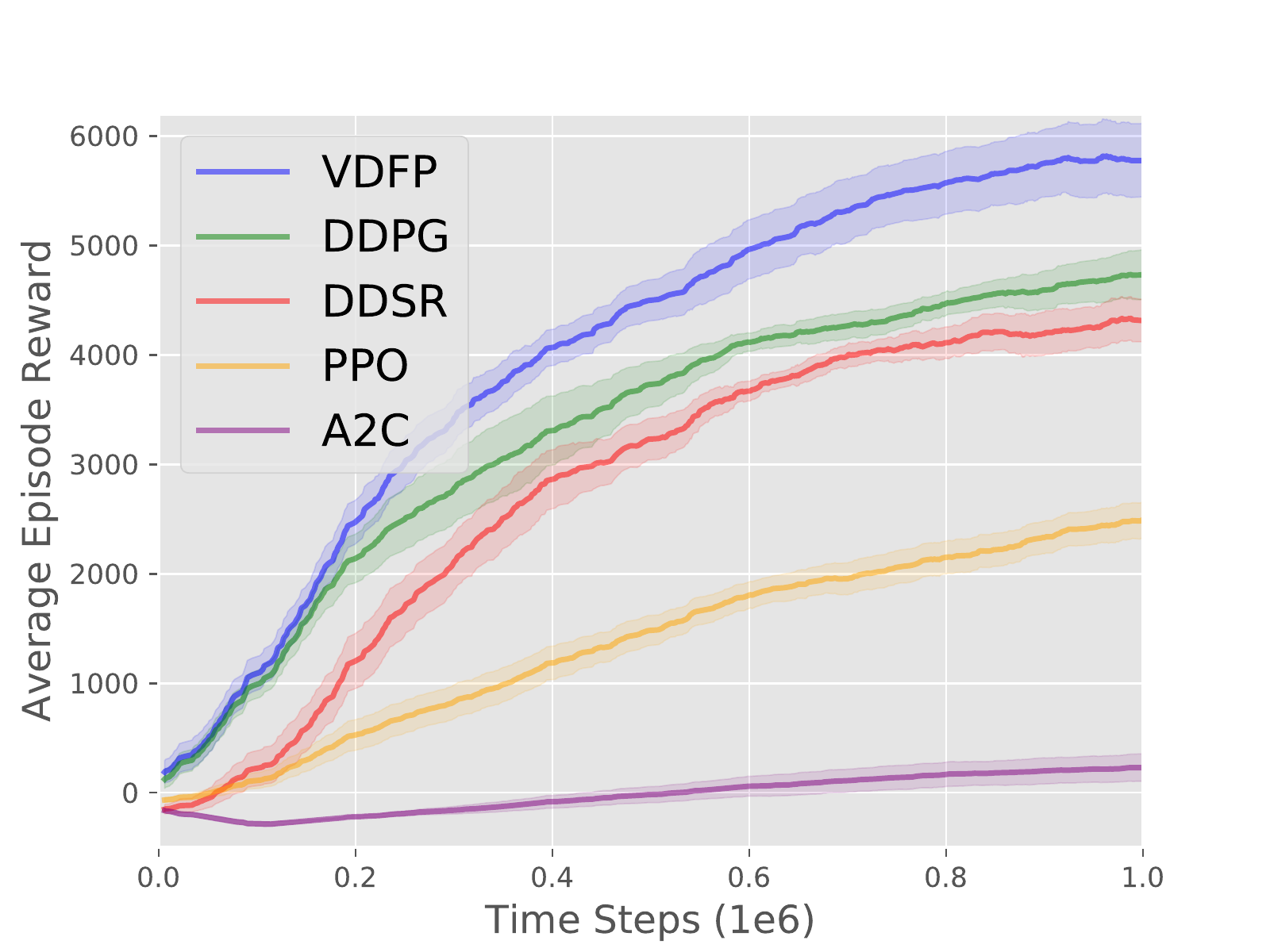}
}
\hspace{-0.8cm}
\subfigure[delay step $d$ = 16]{
\includegraphics[width=0.34\textwidth]{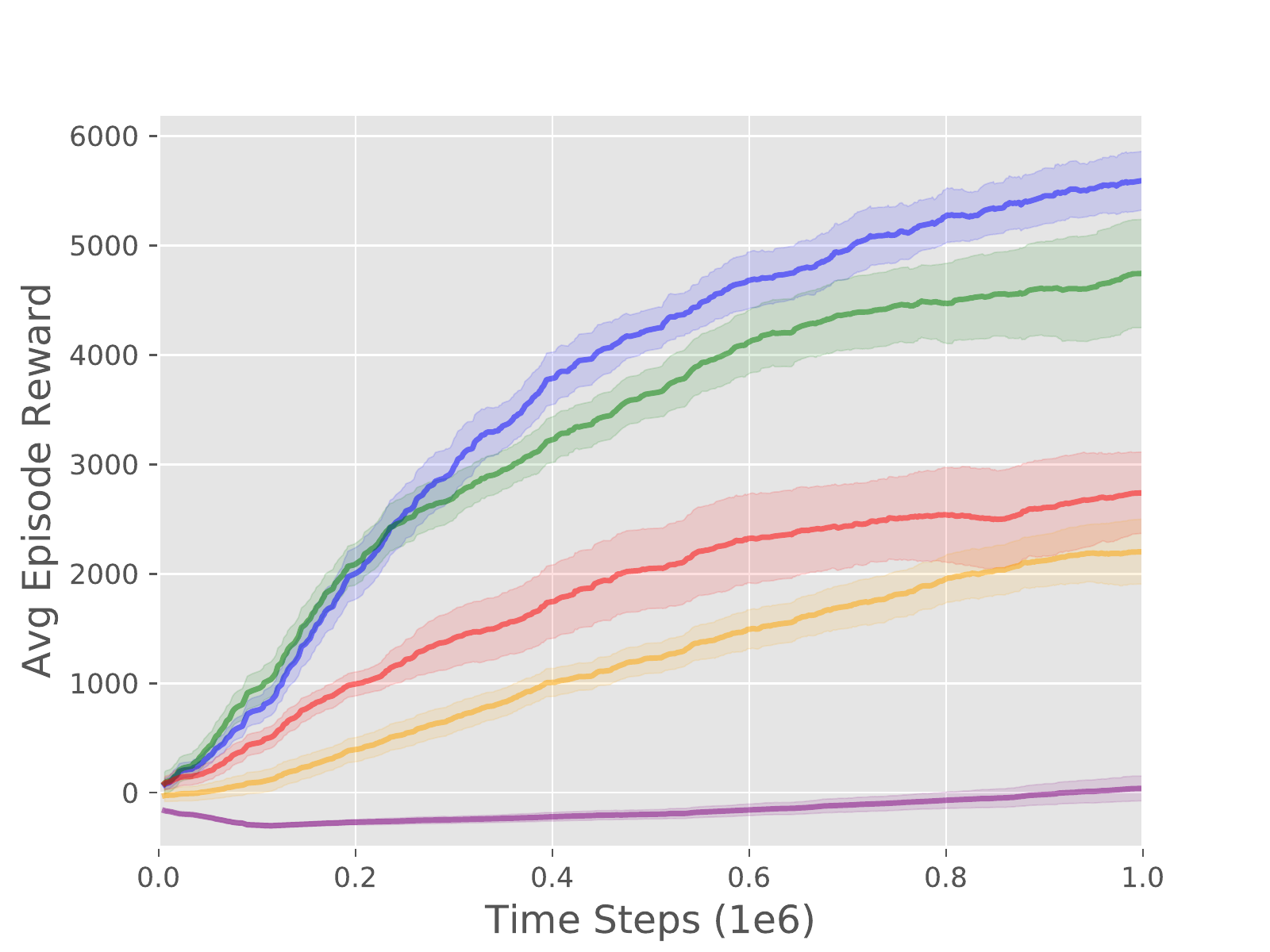}
}
\hspace{-0.8cm}
\subfigure[delay step $d$ = 32]{
\includegraphics[width=0.34\textwidth]{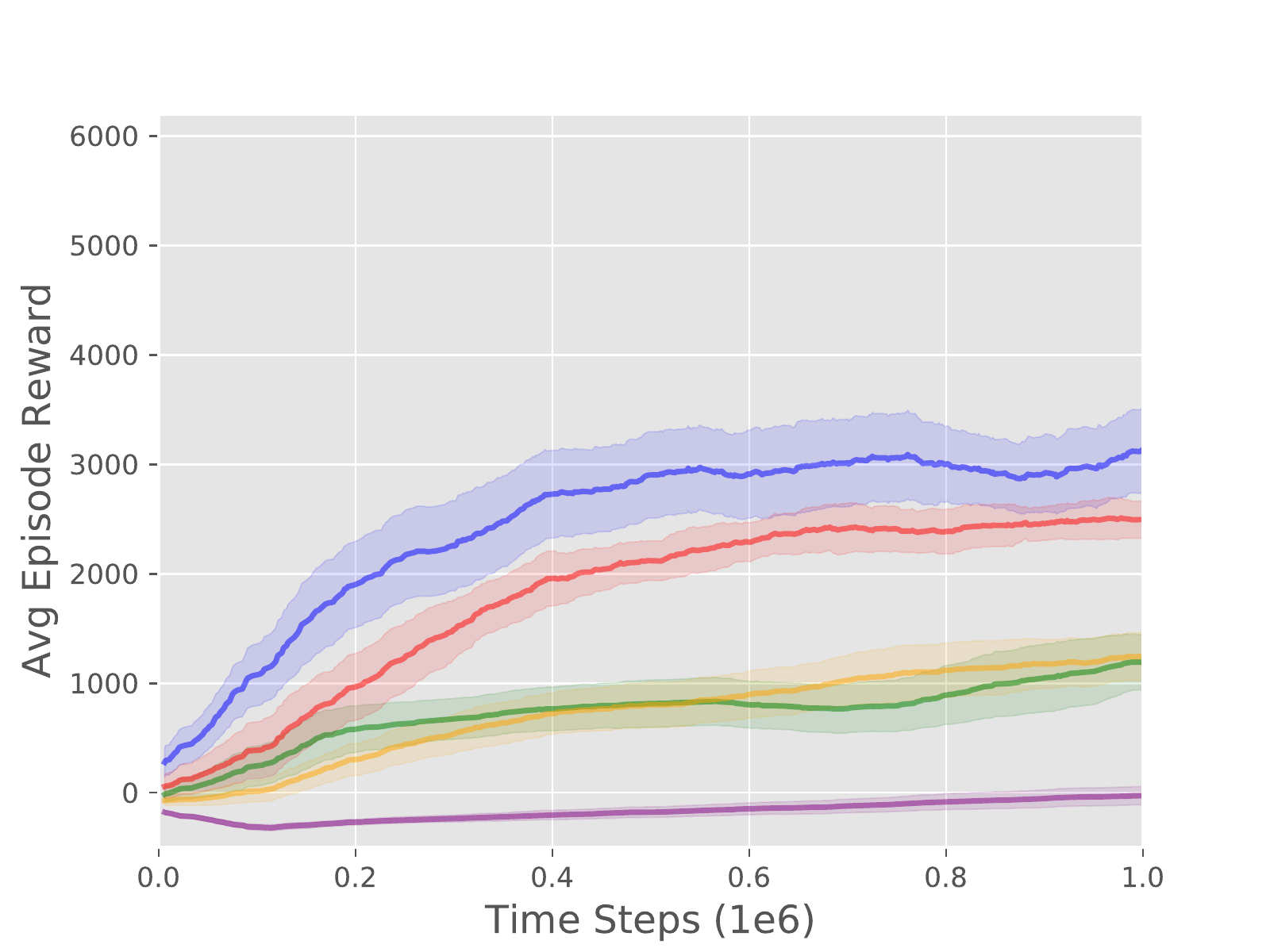}
}
\hspace{-0.8cm}
\subfigure[delay step $d$ = 64]{
\includegraphics[width=0.34\textwidth]{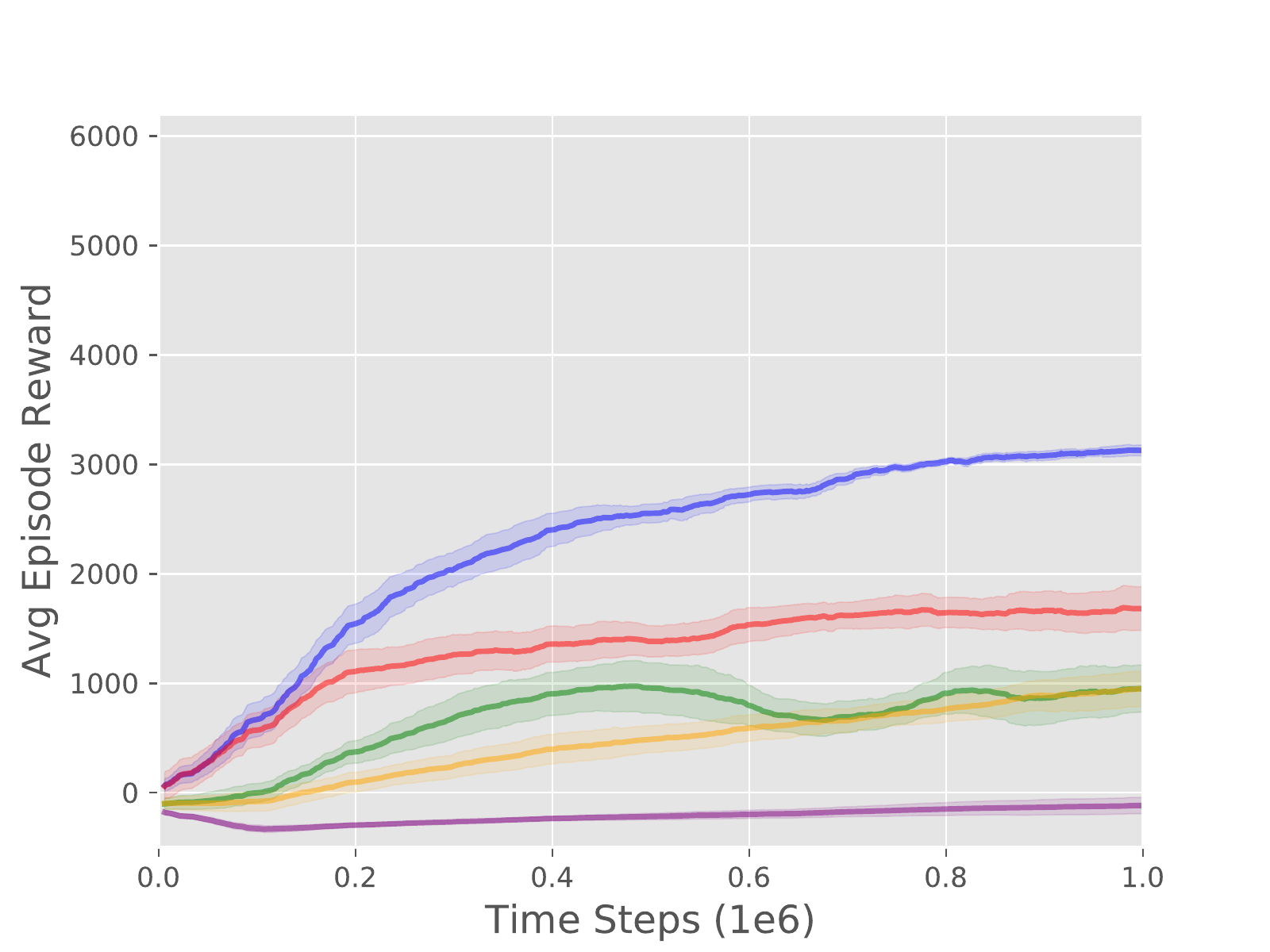}
}
\hspace{-0.8cm}
\subfigure[delay step $d$ = 128]{
\includegraphics[width=0.34\textwidth]{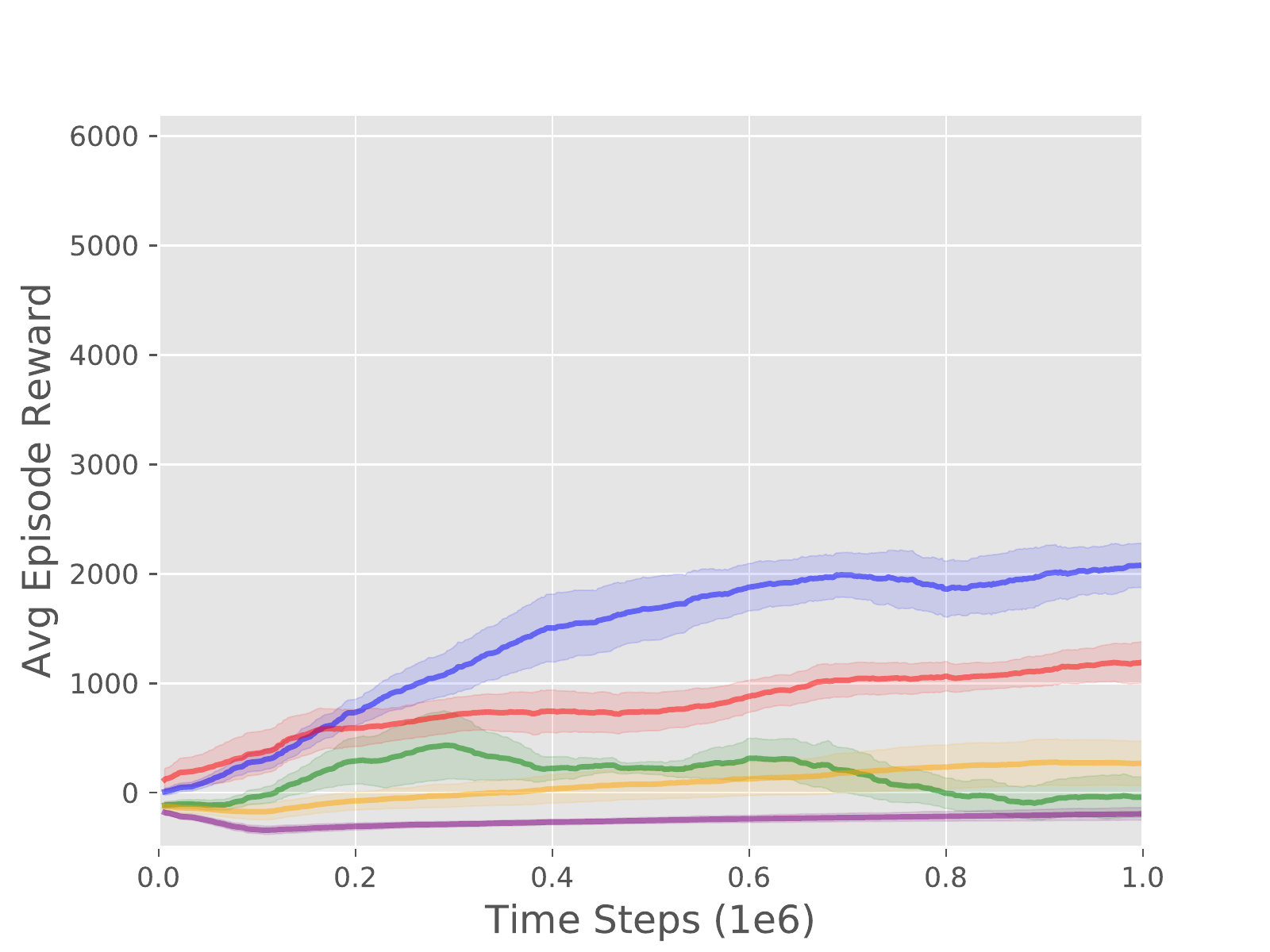}
}

\caption{Learning curves of algorithms in HalfCheetah-v1 under the second delayed reward setting.
Different delay steps are listed from left to right.
The shaded region denotes half a standard deviation of average evaluation over 5 trials.
Results are smoothed over recent 100 episodes.}
%\caption{Cooperative games.}
\label{app_figure:delay_HalfCheetah}
\end{figure}

\begin{figure}
\centering
\subfigure[delay step $d$ = 0]{
\includegraphics[width=0.34\textwidth]{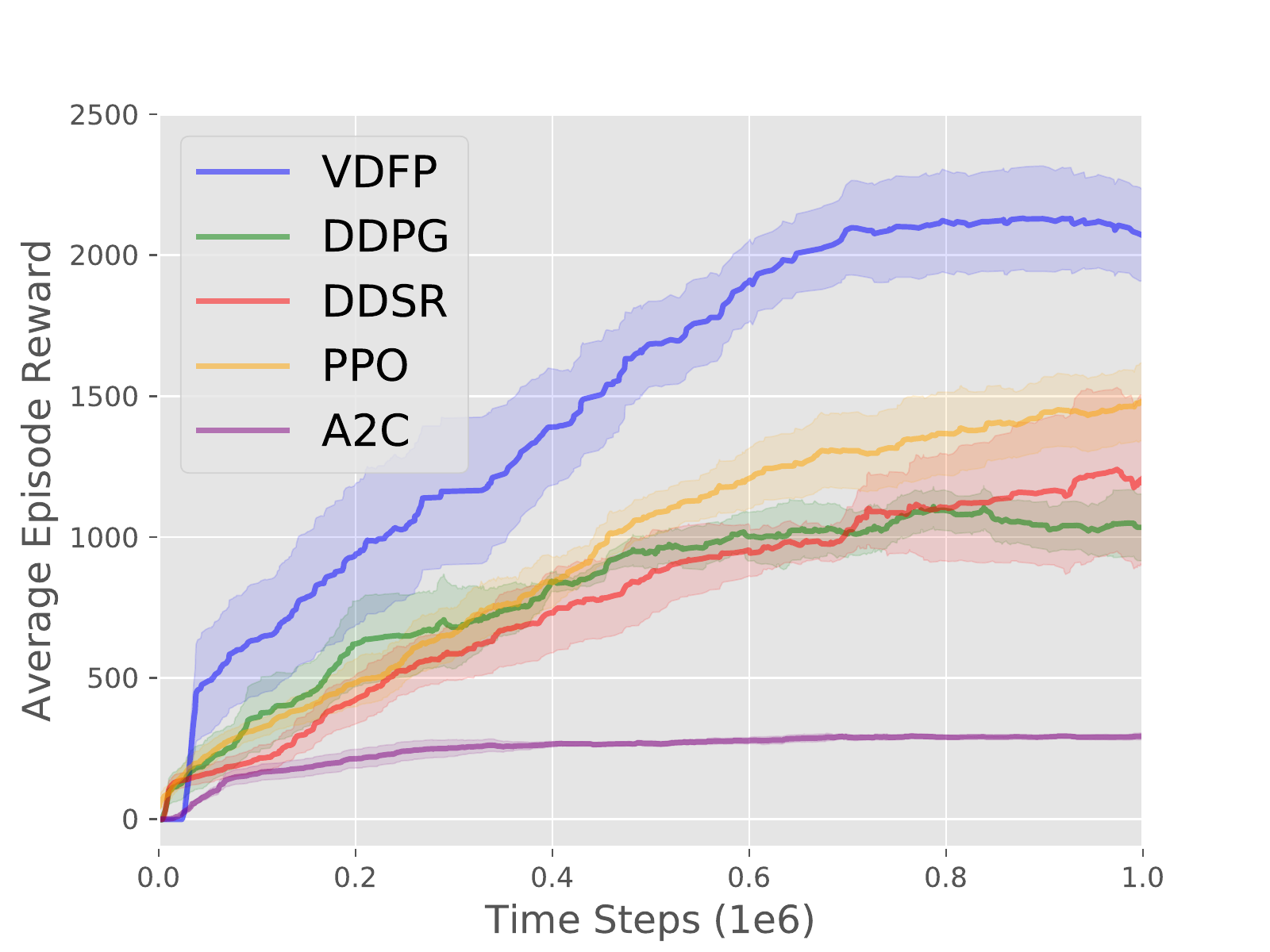}
}
\hspace{-0.8cm}
\subfigure[delay step $d$ = 16]{
\includegraphics[width=0.34\textwidth]{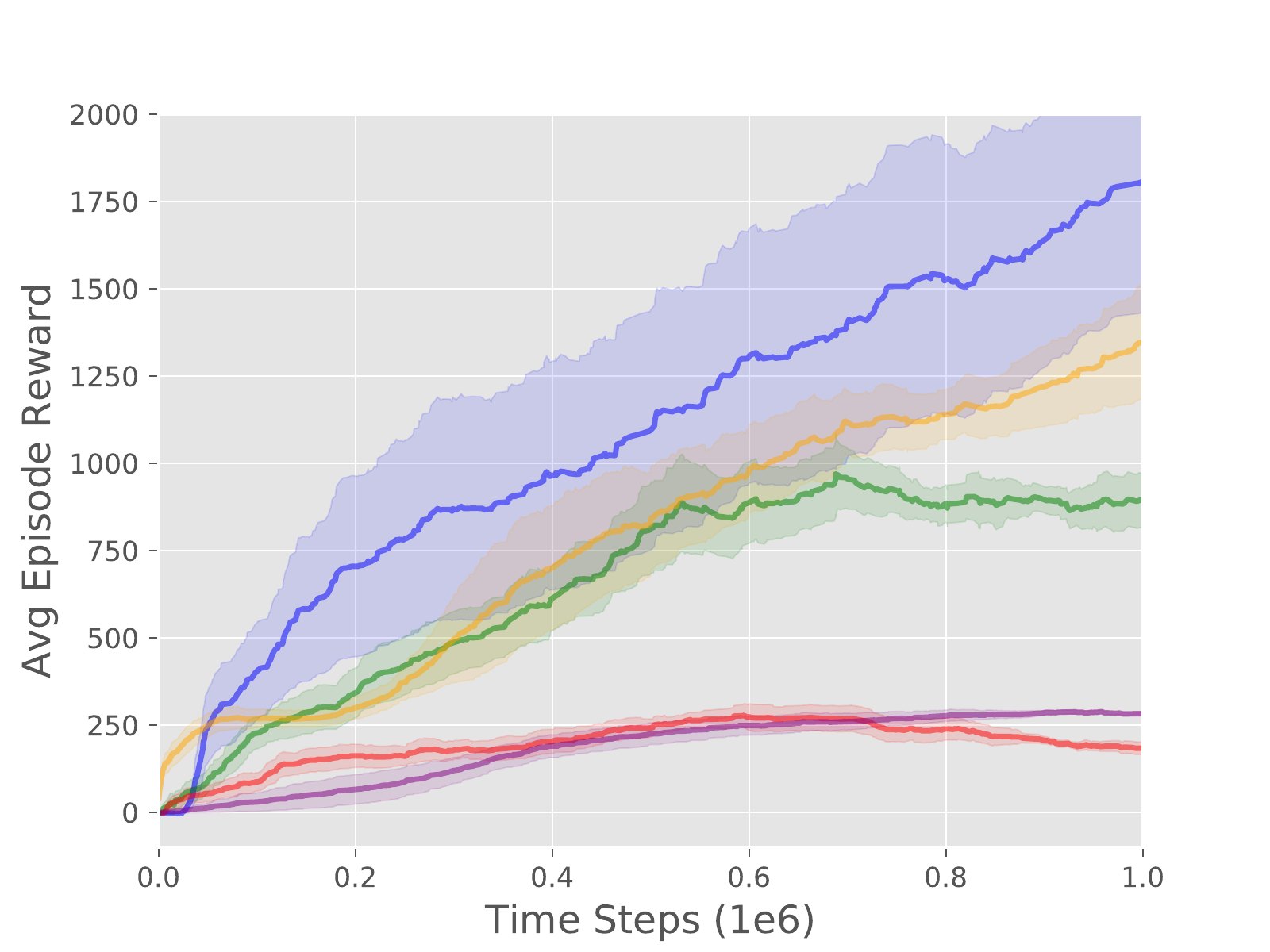}
}
\hspace{-0.8cm}
\subfigure[delay step $d$ = 32]{
\includegraphics[width=0.34\textwidth]{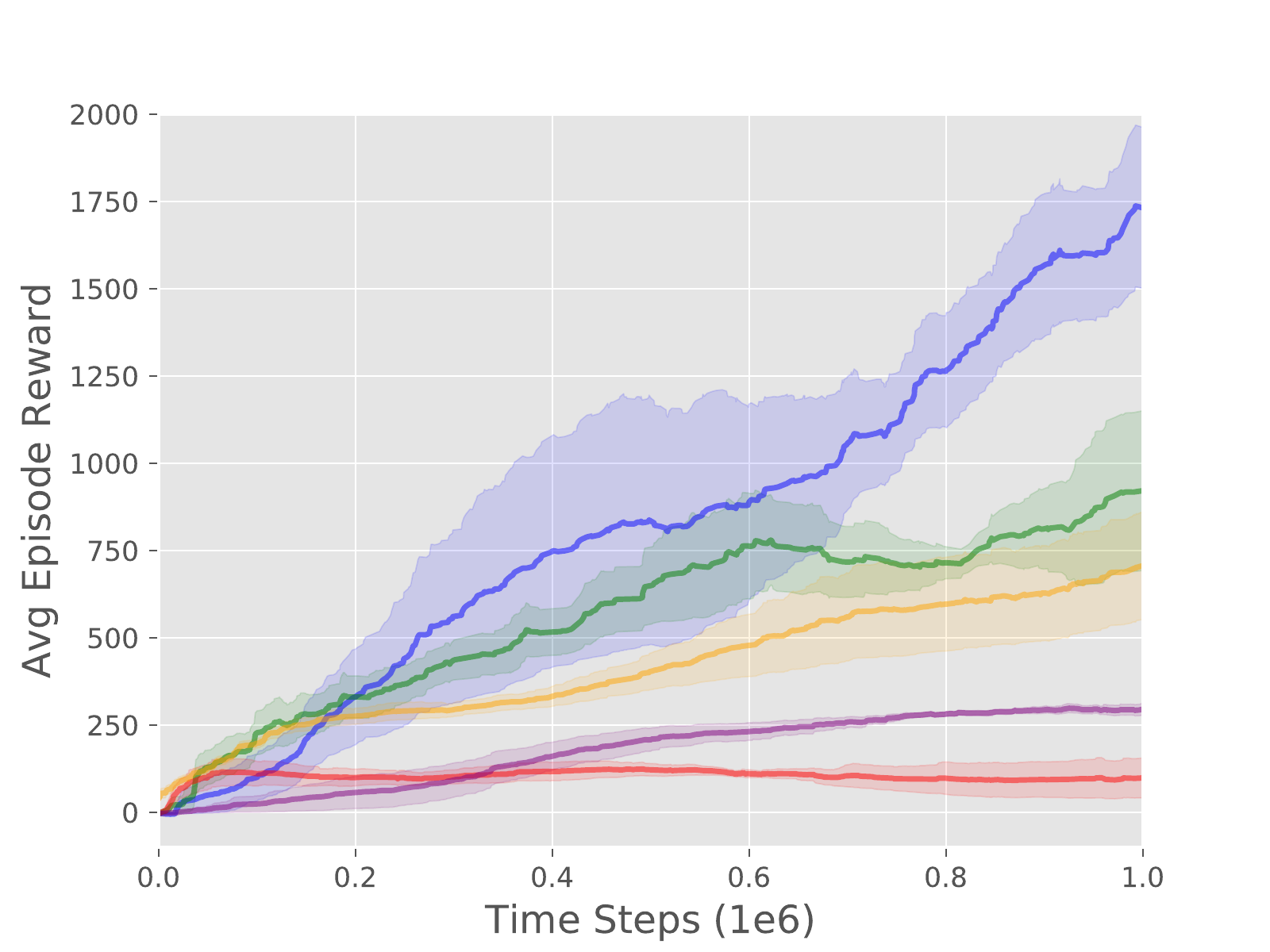}
}
\hspace{-0.8cm}
\subfigure[delay step $d$ = 64]{
\includegraphics[width=0.34\textwidth]{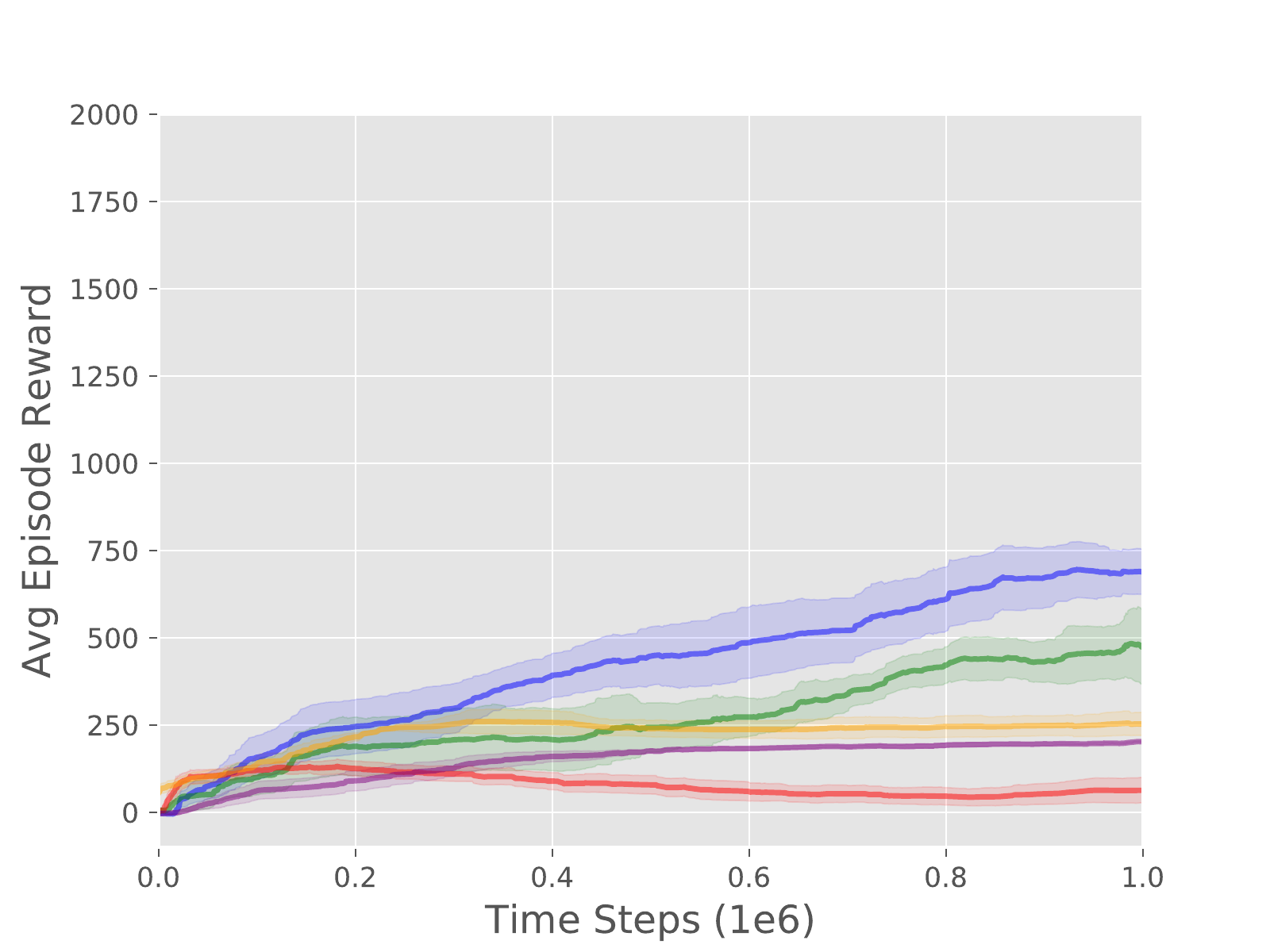}
}
\hspace{-0.8cm}
\subfigure[delay step $d$ = 128]{
\includegraphics[width=0.34\textwidth]{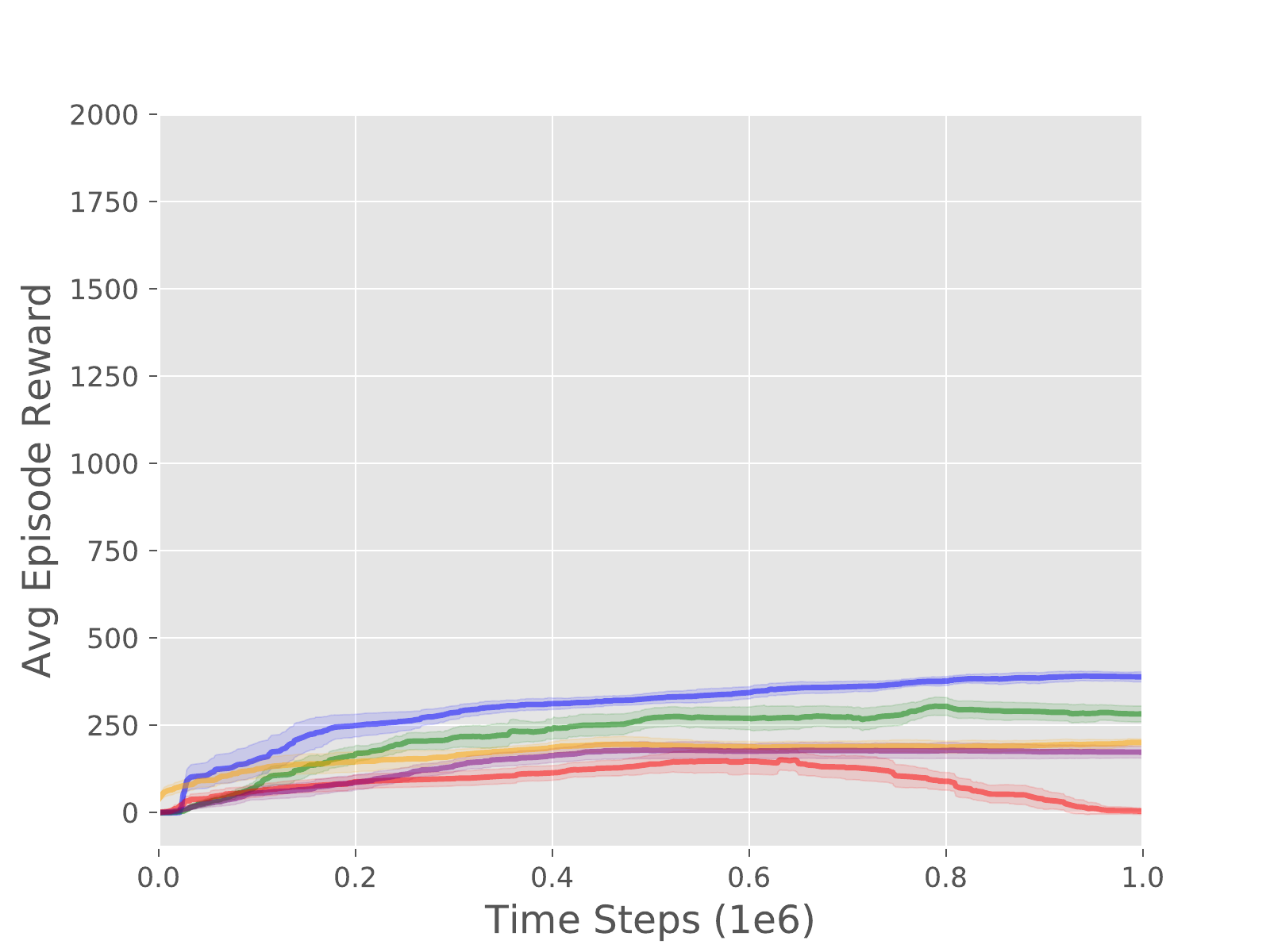}
}

\caption{Learning curves of algorithms in Walker2d-v1 under the second delayed reward setting.
Different delay steps are listed from left to right.
The shaded region denotes half a standard deviation of average evaluation over 5 trials.
Results are smoothed over recent 100 episodes.}
%\caption{Cooperative games.}
\label{app_figure:delay_Walker2d}
\end{figure}

We use the second delayed reward setting to model the representative delay reward in real-world scenarios:
each one-step reward is delayed for certain time steps.
We make a simple modification to MuJoCo tasks to simulate such class of delay reward:
delay the immediate reward of each step by $d$ steps and compensate at the end of episode.
The complete learning curves of algorithms under the second delay reward setting are shown in Figure \ref{app_figure:delay_HalfCheetah} and \ref{app_figure:delay_Walker2d}.
All algorithms gradually degenerate with the increase of delay step $d$.
VDFP consistently outperforms others under all settings, and shows good robustness with delay step $d \le 64$.
%DDSR can hardly learn effective policies due to the failure of its one-step reward model under such delayed reward settings.

\section*{B. Experimental Details}

\subsection*{B.1. Environment Setup}
We conduct our experiments on MuJoCo continuous control tasks in OpenAI gym.
We use the OpenAI gym with version 0.9.1, the mujoco-py with version 0.5.4 and the MuJoCo products with version MJPRO131.
Our codes are implemented with Python 3.6 and Tensorflow 1.8.
%For more details of our code can refer to the \textbf{ReadME\_for\_Code.pdf} and \textbf{SourceCode.zip}.
Our code and raw learning curves are submitted under review, and will be released on GitHub soon.

\subsection*{B.2. Network Structure}

As shown in Table \ref{table:ac_network}, we use a two-layer feed-forward neural network of 200 and 100 hidden units with ReLU activation (except for the output layer) for the actor network for all algorithms, and for the critic network for DDPG, PPO and A2C.
For PPO and A2C, the critic denotes the $V$-network.

\begin{table}[h]
  \caption{Network structures for the actor network and the critic network ($Q$-network or $V$-network).
  }
  \label{table:ac_network}
  \centering
  \scalebox{1.0}{
  \begin{tabular}{ccc}
    \toprule
    Layer & Actor Network ($\pi(s)$) & Critic Network ($Q(s,a)$ or $V(s)$)\\
    \midrule
    Fully Connected & (state dim, 200) & (state dim, 200) \\
    Activation & ReLU & ReLU \\
    \midrule
    Fully Connected & (200, 100) & (action dim + 200, 100) or (200, 100) \\
    Activation & ReLU & ReLU \\
    \midrule
    Fully Connected & (100, action dim) & (100, 1) \\
    Activation & tanh & None \\
    \bottomrule
  \end{tabular}
  }
\end{table}

For DDSR, the factored critic (i.e., $Q$-function) consists of a representation network, a reconstruction network, a SR network and a linear reward vector, as described in the original paper of DSR.
The structure of the factored critic of DDSR is shown in Table \ref{table:ddsr_critic}.

\begin{table}[h]
  \caption{The Network structure for the factored critic network of DDSR, including a representation network, a reconstruction network, a SR network and a reward vector.
  }
  \label{table:ddsr_critic}
  \centering
  \scalebox{1.0}{
  \begin{tabular}{ccc}
    \toprule
    Network & Layer & Structure \\
    \midrule
    Representation Network & Fully Connected & (state dim, 200) \\
     & Activation & ReLU \\
     & Fully Connected & (200, 100) \\
     & Activation & ReLU \\
     & Fully Connected & (100, representation dim) \\
     & Activation & None \\
    \midrule
    Reconstruction Network & Fully Connected & (representation dim, 100) \\
     & Activation & ReLU \\
     & Fully Connected & (100, 200) \\
     & Activation & ReLU \\
     & Fully Connected & (200, state dim) \\
     & Activation & None \\
    \midrule
    SR Network & Fully Connected & (representation dim, 200) \\
     & Activation & ReLU \\
     & Fully Connected & (200, 100) \\
     & Activation & ReLU \\
     & Fully Connected & (100, representation dim) \\
     & Activation & None \\
    \midrule
    Linear Reward Vector & Fully Connected (not use bias) & (representation dim, 1) \\
     & Activation & None \\
    \bottomrule
  \end{tabular}
  }
\end{table}

For VDFP, the decomposed critic (i.e., $Q$-function) consists of a convolutional representation network $f^{\rm CNN}$, a linear trajectory return network $U^{\rm Linear}$ and a conditional VAE (an encoder network and a decoder network).
The structure of the decomposed critic of VDFP is shown in Table \ref{table:vdfp_critic}.

\begin{table}[h]
  \caption{The Network structure for the factored critic network of VDFP, including a convolutional representation network $f^{\rm CNN}$, a linear trajectory return network $U^{\rm Linear}$ and a conditional VAE (an encoder network and a decoder network).
  }
  \label{table:vdfp_critic}
  \centering
  \scalebox{0.85}{
  \begin{tabular}{ccc}
    \toprule
    Network & Layer (Name) & Sturcture \\
    \midrule
    Representation Network & Convolutional & filters with height $\in \{1,2,4,8,16,32,64\}$ \\
    $f^{\rm CNN}(\tau_{t:t+k})$ & & of numbers $ \{20, 20, 10, 10, 5, 5, 5\}$ \\
     & Activation & ReLU \\
     & Pooling (concat) & Maxpooling \& Concatenation \\
     & Fully Connected (highway) & (filter num, filter num) \\
     & Joint & Sigmoid(highway) $\cdot$ ReLU(highway)\\
     & & + (1 - Sigmoid(highway)) $\cdot$ ReLU(concat)\\
     & Dropout & Dropout(drop\_prob = 0.2) \\
     & Fully Connected & (filter num, representation dim) \\
     & Activation & None \\
    \midrule
    Conditional Encoder Network & Fully Connected (main) & (representation dim, 400) \\
    $q_{\phi}(z_t|m_{t:t+k}, s_t, a_t)$ & Fully Connected (encoding) & (state dim + action dim, 400) \\
     & Pairwise-Product & Sigmoid(encoding) $\cdot$ ReLU(main) \\
     & Fully Connected & (400, 200) \\
     & Activation & ReLU \\
     & Fully Connected (mean) & (200, $z$ dim) \\
     & Activation & None \\
     & Fully Connected (log\_std) & (200, $z$ dim) \\
     & Activation & None \\
    \midrule
    Conditional Decoder Network & Fully Connected (latent) & (z dim, 200) \\
    $p_{\varphi}(m_{t:t+k}| z_t, s_t, a_t)$ & Fully Connected (decoding) & (state dim + action dim, 200) \\
     & Pairwise-Product & Sigmoid(decoding) $\cdot$ ReLU(latent) \\
     & Fully Connected & (200, 400) \\
     & Activation & ReLU \\
     & Fully Connected (reconstruction) & (400, representation dim) \\
     & Activation & None \\
    \midrule
    Trajectory Return Network & Fully Connected & (representation dim, 1) \\
    $U^{\rm Linear}(m_{t:t+k})$ & Activation & None \\
    \bottomrule
  \end{tabular}
  }
\end{table}

For VDFP\_MLP, we also use a two-layer feed-forward neural network of 200 and 100 hidden units with ReLU activation (except for the output layer) for $P^{\rm MLP}$.
For VDFP\_LSTM, we use one LSTM layer with 100 units to replace the convolutional layer (along with maxpooling layer) as described in Table \ref{table:vdfp_critic}.
For VDFP\_Concat, we concatenate the state, action and the representation (or latent variable) rather than a pairwise-product structure.
For VDFP\_ReLU, we add an ReLU-activated fully-connected layer with 50 units in front of the linear layer for trajectory return model.

\subsection*{B.3. Hyperparameter}

For all our experiments, we use the raw observation and reward from the environment and no normalization or scaling are used.
No regularization is used for the actor and the critic in all algorithms.
Table \ref{table:hyperparameter} shows the common hyperparamters of algorithms used in all our experiments.
For VDFP and DDSR, critic learning rate denotes the learning rate of the conditional VAE and the successor representation model respectively.
Return (reward) model learning rate denotes the learning rate of the return model (along with the representation model) for VDFP and the learning rate of the immediate reward vector for DDSR.

\begin{table}[h]
  \caption{A comparison of common hyperparameter choices of algorithms.
  We use `-' to denote the `not applicable' situation.
  }
  \label{table:hyperparameter}
  \centering
  \scalebox{1.0}{
  \begin{tabular}{c|c|cc|cc}
    \toprule
    Hyperparameter & VDFP & DDSR & DDPG & PPO & A2C\\
    \midrule
    Actor Learning Rate & 2.5$\cdot$10$^{-4}$ & 2.5$\cdot$10$^{-4}$ & 10$^{-4}$ & 10$^{-4}$ & 10$^{-4}$ \\
    Critic (VAE, SR) Learning Rate & 10$^{-3}$ & 10$^{-3}$ & 10$^{-3}$ & 10$^{-3}$ & 10$^{-3}$ \\
    Return (Reward) Model Learning Rate & 5$\cdot$10$^{-4}$ & 5$\cdot$10$^{-4}$ & - & - & - \\
    \midrule
    Discount Factor & 0.99 & 0.99 & 0.99 & 0.99 & 0.99 \\
    Optimizer & Adam & Adam & Adam & Adam & Adam\\
    Target Update Rate & - & 10$^{-3}$ & 10$^{-3}$ & - & -\\
    Exploration Policy & $\mathcal{N}(0, 0.1)$ & $\mathcal{N}(0, 0.1)$ & $\mathcal{N}(0, 0.1)$ & None & None\\
    Batch Size & 64 & 64 & 64 & 256 & 256 \\
    Buffer Size & 10$^{5}$ & 10$^{5}$ & 10$^{5}$ & - & - \\
    Actor Epoch & - & - & - & 10 & 10 \\
    Critic Epoch & - & - & - & 10 & 10 \\
    \bottomrule
  \end{tabular}
  }
\end{table}

\subsection*{B.4. Additional Implementation Details}

For DDPG, the actor network and the critic network is updated every 1 time step.
We implement the DDSR based on the DDPG algorithm, by replacing the original critic of DDPG with the factored $Q$-function as described in the DSR paper.
The actor, along with all the networks described in Table \ref{table:ddsr_critic} are updated every 1 time step.
The representation dimension is set to 100.
Before the training of DDPG and DDSR, we run 10000 time steps for experience collection, which are also counted in the total time steps.

For PPO and A2C, we use Generalized Advantage Estimation with $\lambda = 0.95$ for stable policy gradient.
The clip range of PPO algorithm is set to 0.2.
The actor network and the critic network are updated every $2$ and $5$ episodes for HalfCheetah-v1 and Walker2d-v1 respectively, with the epoches and batch sizes described in Table \ref{table:hyperparameter}.

For VDFP, we set the KL weight $\beta$ as 1000 and the clip value $c$ as 0.2.
The latent variable dimension ($z$ dim) is set to 50 and the representation dimension is set to 100.
We collect trajectories experiences in the first 5000 time steps and then pre-train the trajectory model (along with the representation model) and the conditional VAE for 15000 time steps, after which we start the training of the actor.
All the time steps above are counted into the total time steps for a fair comparison.
The trajectory return model (along with the representation model) is trained every 10 time steps for the pre-train process and is trained every 50 time steps in the rest of training, which already ensures a good performance in our experiments.
The actor network and the conditional VAE are trained every 1 time step.

We consider a max trajectory length $l$ for VDFP in our experiments.
For example, a max length $l$ = 256 can be considered to correspond a discounted factor $\gamma = 0.99$ as $0.99^{256} \approx 0.076$.
In practice, for a max length $l > 64$, we add an additional fully-connected layer with ReLU activation before the convolutional representation model $f^{\rm CNN}$, to aggregate the trajectory into the length of 64.
This is used for the purpose of reducing the time cost and accelerating the training of the convolutional representation model as described in Table \ref{table:vdfp_critic}.
For example, we feed every 4 state-action pairs of a trajectory with length 256 into the aggregation layer to obtain an aggregated trajectory with the length 64, and then feed the aggregated trajectory to $f^{\rm CNN}$ for a trajectory representation.

\section*{C. Complete Formulation for the conditional VAE}

A variational auto-encoder (VAE) is a generative model which aims to maximize the marginal log-likelihood $\log p(X) = \sum_{i=1}^N \log p(x_i)$ where $X = \{x_1, \dots ,x_N\}$, the dataset.
While computing the marginal likelihood is intractable in nature, it is a common choice to train the VAE through optimizing the variational lower bound:
\begin{equation}
    \log p(X) \ge \mathbb{E}_{q(X|z)}[\log p(X|z)] + D_{\rm KL} \big(q(z|X) \| p(z) \big),
\end{equation}
where $p(z)$ is chosen a prior, generally the multivariate normal distribution $\mathcal{N}(0,I)$.

In our paper, we use a conditional VAE to model the latent distribution of the representation of  future trajectory conditioned on the state and action, under certain policy.
Thus, the true parameters of the distribution, denoted as $\varphi^*$, maximize the conditional log-likelihood as follows:
\begin{equation}
    \varphi^* = \arg \max_{\varphi} \sum_{i=1}^N \log p^{\pi}_{\varphi}(m_{t:t+k}|s_t,a_t).
\end{equation}
The likelihood can be calculated with the prior distribution of latent variable $z$:
\begin{equation}
    p^{\pi}_{\varphi}(m_{t:t+k}|s_t,a_t) = \int p^{\pi}_{\varphi}(m_{t:t+k}|z_t, s_t, a_t) p^{\pi}_{\varphi}(z_t) {\rm d}z.
\end{equation}
Since the prior distribution of latent variable $z$ is not easy to compute, an approximation of posterior distribution $q^{\pi}_{\phi}(z_t|m_{t:t+k},s_t,a_t)$ with parameterized $\phi$ is introduced.

The variational lower bound of such a conditional VAE can be obtained as follows, superscripts and subscripts are omitted for clarity:
\begin{equation}
\begin{aligned}
        D_{\rm KL} & \big(q_{\phi}(z|m,s,a) \| p_{\varphi}(z|m,s,a) \big)  \\
        & = \int q_{\phi}(z|m,s,a) \log \frac{q_{\phi}(z|m,s,a)}{p_{\varphi}(z|m,s,a)} {\rm d}z \\
        & = \int q_{\phi}(z|m,s,a) \log \frac{q_{\phi}(z|m,s,a) p_{\varphi}(m|s,a)}{p_{\varphi}(z,m|s,a)} {\rm d}z \\
        & = \int q_{\phi}(z|m,s,a) \big[ \log p_{\varphi}(m|s,a) +  \log \frac{q_{\phi}(z|m,s,a) }{p_{\varphi}(z,m|s,a)} \big]{\rm d}z \\
        & = \log p_{\varphi}(m|s,a) + \int q_{\phi}(z|m,s,a)  \log \frac{q_{\phi}(z|m,s,a) }{p_{\varphi}(z,m|s,a)} {\rm d}z \\
        & = \log p_{\varphi}(m|s,a) + \int q_{\phi}(z|m,s,a)  \log \frac{q_{\phi}(z|m,s,a) }{p_{\varphi}(m|z,s,a) p_{\varphi}(z|s,a)} {\rm d}z \\
        & = \log p_{\varphi}(m|s,a) + \int q_{\phi}(z|m,s,a) \big[ \log \frac{q_{\phi}(z|m,s,a) }{p_{\varphi}(z|s,a)} - \log p_{\varphi}(m|z,s,a) \big] {\rm d}z \\
        & = \log p_{\varphi}(m|s,a) + D_{\rm KL} \big( q_{\phi}(z|m,s,a) \| p_{\varphi}(z|s,a) \big) - \mathbb{E}_{z \sim q_{\phi}(z|m,s,a)} \log p_{\varphi}(m|z,s,a). \\
\end{aligned}
\end{equation}
Re-arrange the above equation,
\begin{equation}
\begin{aligned}
    \log p_{\varphi}(m|s,a) - & D_{\rm KL} \big(q_{\phi}(z|m,s,a) \| p_{\varphi}(z|m,s,a) \big) \\
    & =  \mathbb{E}_{z \sim q_{\phi}(z|m,s,a)} \log p_{\varphi}(m|z,s,a)
    - D_{\rm KL} \big( q_{\phi}(z|m,s,a) \| p_{\varphi}(z|s,a) \big). \\
\end{aligned}
\end{equation}
Since the KL divergence is none-negative, we obtain the variational lower bound for the conditional VAE:
\begin{equation}
    \log p_{\varphi}(m|s,a) \ge \log p_{\varphi}(m|s,a) - D_{\rm KL} \big(q_{\phi}(z|m,s,a) \| p_{\varphi}(z|m,s,a) \big) = - \mathcal{L}^{\rm VAE},
\end{equation}
Thus, we can obtain the optimal parameters through optimizing the equation below:
\begin{equation}
\label{equation:elbo}
\begin{aligned}
    \mathcal{L}^{\rm VAE} = - \mathbb{E}_{z \sim q_{\phi}(z|m,s,a)} \log p_{\varphi}(m|z,s,a) &
    + D_{\rm KL} \big( q_{\phi}(z|m,s,a) \| p_{\varphi}(z|s,a) \big), \\
    \phi^*, \varphi^* = & \arg \min_{\phi,\varphi} \mathcal{L}^{\rm VAE}.
\end{aligned}
\end{equation}

$q_{\phi}(z|m,s,a)$ is the conditional variational encoder and the $\log p_{\varphi}(m|z,s,a)$ is the conditional variational decoder.

In our paper, we implement the encoder and decoder with deep neural networks.
The encoder takes the trajectory representation and state-action pair as input and output a Gaussian distribution with mean $\mu_t$ and standard deviant $\sigma_t$, from which a latent variable is sampled and then feed into the decoder for the reconstruction of the trajectory representation.
Thus, we train the conditional VAE with respect to the variational lower bound (Equation \ref{equation:elbo}), in the following form:
\begin{equation}
    \mathcal{L}^{\rm VAE}(\phi, \varphi) = \mathbb{E}_{\tau_{t:t+k} \sim \mathcal{D}} \Big[ \|m_{t:t+k} - \tilde{m}_{t:t+k}\|_2^2
    + D_{\rm KL} \big( \mathcal{N}(\mu_t,\sigma_t) \| \mathcal{N}(0,I) \big) \Big].
\end{equation}

\end{document}